\documentclass{article}

\usepackage{PRIMEarxiv}

\usepackage[utf8]{inputenc} 
\usepackage[T1]{fontenc}    
\usepackage{hyperref}       
\usepackage{url}            
\usepackage{booktabs}       
\usepackage{amsfonts}       
\usepackage{amsmath}
\usepackage{nicefrac}       
\usepackage{microtype}      
\usepackage{fancyhdr}       
\usepackage{graphicx}       
\usepackage{subcaption}
\usepackage{tabularx}
\usepackage{caption}
\usepackage{placeins}
\usepackage{float}
\usepackage{makecell}
\usepackage{threeparttable}
\usepackage{multirow}
\usepackage{array}
\usepackage[numbers,sort&compress]{natbib}

\graphicspath{{./}{media/}}

\title{Making AI-Generated Feedback Matter: A Large-Scale Study of Feedback Workflows and Student Enactment}

\author{
  Omar Alsaiari\thanks{Corresponding author.} \\
  The University of Queensland \\
 Brisbane, Australia \\
  \texttt{o.alsaiari@uq.edu.au} \\
  \And
  Nilufar Baghaei \\
  The University of Queensland \\
 Brisbane, Australia \\
  \texttt{n.baghaei@uq.edu.au} \\
  \AND
  Jason M. Lodge \\
  The University of Queensland \\
  Brisbane, Australia \\
  \texttt{jason.lodge@uq.edu.au} \\
  \And
  Naomi Winstone \\
  University of Surrey \\
  Guildford, UK \\
  \texttt{n.winstone@surrey.ac.uk} \\ 
  \AND
 Dragan Ga\v{s}evi\'{c} \\
  The University of Hong Kong \\
  Hong Kong, China \\
  \texttt{dgasevic@hku.hk} \\
  \And
  Hassan Khosravi \\
  The University of Queensland \\
  Brisbane, Australia \\
  \texttt{h.khosravi@uq.edu.au} \\
}

\begin{document}
\maketitle

\begin{abstract}
Feedback processes strongly influence student learning, yet realising their educational potential requires addressing two distinct barriers: providing high-quality, timely, and individualised feedback comments at scale, and supporting students to interpret, evaluate, and act on those comments productively. Generative AI offers a credible means of addressing the provision barrier, with growing evidence that AI-generated feedback comments can be pedagogically comparable to educator-generated feedback comments. However, students' uptake remains limited, making productive feedback use a critical constraint on the educational value of AI-generated feedback comments. We conducted a large-scale quasi-experimental sequential cohort study comparing three AI-mediated feedback workflows across 13,037 students, 51,296 student-authored resources, and 70 course offerings. In \textit{Directed Feedback} (n = 3{,}723), students received AI-generated feedback comments without structured support. In \textit{Self-Directed Feedback} (n = 3{,}951), students could initiate optional AI-supported dialogue. In \textit{Enacted Feedback} (n = 5{,}363), the workflow prompted students to select feedback suggestions, evaluate their relevance, and engage in targeted AI-supported dialogue anchored to those selections. \textit{Enacted Feedback} was associated with significantly higher uptake of AI-generated feedback comments, with an estimated probability of 26.2\%, compared with 14.1\% for \textit{Directed Feedback} and 0.1\% for \textit{Self-Directed Feedback}. It was also associated with significantly higher self-assessment confidence and submitted-work quality than both comparison conditions. These findings suggest that the educational value of AI-generated feedback depends not only on the quality of feedback comments, but also on AI-mediated feedback workflows that actively structure students' enactment of feedback literacy processes. The paper discusses implications for the design of AI feedback systems that position learners not as passive recipients of feedback comments, but as active participants in processes of judgement, dialogue, and improvement.
\end{abstract}


\section{Introduction}

Feedback processes are widely recognised as being among the most powerful influences on student learning and academic achievement \citep{hattie2007power, nicol2021power}. When feedback processes are effective, they provide learners with information about the gap between their current performance and desired standards, stimulate internal comparison processes, and support the kinds of evaluative and revision activity that can drive improvement \citep{nicol2021power, carless2018development}. The capacity of feedback processes to accelerate learning, support self-direction, and develop students' capacity for independent judgement has positioned them as a central concern of assessment research and higher education policy alike \citep{winstone2017supporting, boud2013rethinking}. Yet despite this theoretical promise, the educational potential of feedback processes is persistently limited by difficulties in both the provision of high-quality feedback comments and students' productive engagement with them.

These two difficulties, one concerning the provision of feedback comments and the other concerning students' capacity to engage with them productively, are closely intertwined in practice yet conceptually distinct, and each warrants separate consideration. The first challenge concerns provision: generating high-quality, timely, and individualised feedback comments at the scale demanded by contemporary higher education is a structural and pedagogical problem that has long constrained feedback practice. The provision of feedback comments requires subject expertise, familiarity with individual students' work, and the time and capacity to translate that expertise into actionable comments, all of which are chronically scarce, particularly in large-enrolment courses where educator workloads are high and contact time is limited \citep{paris2022instructors,boud2013rethinking}. However, the provision challenge is not only a matter of scale or workload. The quality of feedback comments is also shaped by teachers' feedback literacy, quality assurance expectations, and disciplinary norms, which influence what counts as useful, appropriate, and actionable feedback comments in different contexts \citep{haughney2020quality, deneen2023connecting,esterhazy2020counts}. As a consequence, feedback comments in higher education are frequently delayed, generic, or insufficiently detailed to support meaningful revision \citep{winstone2019designing}. The second challenge concerns engagement: feedback is educationally valuable only when students actively use it to compare their current performance with relevant criteria and desired standards, thereby generating internal feedback \citep{nicol2021power}. This distinction between the provision of feedback comments and feedback use, where learners are expected to actively interpret, select, and act upon feedback comments, is the central tension animating contemporary research on assessment and feedback \citep{winstone2017supporting, carless2018development, Winstone03072023}.

The emergence of generative artificial intelligence (GenAI) as a feedback 
agent has introduced a credible means of addressing the first of the above two 
challenges at scale. In this study, the term \textit{AI-generated feedback comments} refers specifically to feedback comments produced by GenAI models. GenAI-powered systems can now make real-time, individualised feedback comments available at a speed and volume that far exceed what educators can 
provide, and there is accumulating evidence that such comments are 
pedagogically sound. Recent empirical studies indicate that AI-generated feedback comments can, under some conditions, match or approach educator-generated feedback comments in pedagogical quality and in supporting students' revision outcomes \citep{NAZARETSKY2026100533, MUNOZMUNOZ2025103805, alnemrat2025ai}. Whilst AI-generated feedback comments may be similar in form and quality to educator-generated comments, the literature shows a clear preference for educator-generated feedback comments in terms of trust \citep{nazaretsky2026gives}. In higher education systems where student evaluations of teaching are high-stakes quality metrics, universities may therefore be reluctant to rely more heavily on AI-generated feedback comments at the expense of educator-generated comments, for fear of negatively affecting the student experience.
Yet despite these technological advances, students' uptake and productive use of AI-generated feedback comments remain consistently limited \citep{pozdniakov2026}. This indicates that addressing the provision challenge does not, by itself, resolve the challenge of feedback use.

Contemporary feedback scholarship has increasingly turned to the concept of \textit{feedback literacy} to explain and address the persistent gap between the provision of feedback comments and feedback use \citep{carless2018development}. Feedback literacy foregrounds student agency as an important mechanism through which feedback becomes educationally meaningful, and 
positions productive feedback engagement not as a natural response to good 
feedback, but as a learned capability that must be actively developed \citep{carless2018development, molloy2020developing, carless2022teacher}. Critically, feedback literacy is not a fixed attribute that students either 
possess or lack, but a developmental capacity that emerges through 
participation in purposefully structured feedback encounters that require 
active engagement with evaluative processes rather than passive receipt of 
commentary \citep{ajjawi2018conceptualising, winstone2019designing, 
carless2018development}. Central to this development is evaluative judgement \citep{tai2018developing}, which in the present context supports students in assessing the relevance and applicability of AI-generated suggestions, while translating those judgements into revision involves broader self-regulatory processes. This capacity takes on particular 
importance when AI-generated feedback comments may appear authoritative 
and convincing even when pedagogically limited \citep{bearman2024developing}. 
If feedback literacy develops through participation in designed encounters, 
then the workflow structure of those encounters constitutes the primary 
lever for intervention. Yet this structural dimension remains insufficiently addressed in research on AI-generated feedback comments and in the development of AI-mediated feedback systems, which continue to focus predominantly on generating feedback comments rather than on enabling students to interpret, synthesise, and act upon them.

Despite growing recognition that feedback literacy develops through active participation in evaluative processes, AI-generated feedback comments are still predominantly presented through static workflows. Feedback comments are generated and presented, and the pedagogical process is treated as complete. Whether and how students engage with those comments is left largely to their individual discretion and capability. This is reflected in the empirical literature, 
which has predominantly evaluated AI-generated feedback comments as static products, comparing their perceived quality or outcomes with human or peer-generated alternatives, rather than examining how the design of the feedback encounter itself shapes engagement \citep{moore2024harnessing, zhan2025generative}. Where workflow variations have been studied, the mechanisms through which different design features operate have rarely been systematically decomposed. In particular, it remains unclear whether providing access to dialogic AI support is sufficient to improve engagement, or whether structured guidance through processes of selection, evaluation, and targeted interaction is necessary
\citep{guo2025peer}. This represents a substantive empirical gap: if feedback literacy requires students to actively process, evaluate, and act on feedback, then the workflow structure of the feedback encounter is the primary site of intervention, and yet its systematic variation across AI-supported conditions has rarely been the direct focus of empirical investigation.

The present study examines how different AI-mediated feedback workflows shape students' behavioural engagement by comparing three theoretically distinct workflows implemented at scale within an authentic higher education environment \citep{liu2025ai, ziqi2024l2}. Critically, the three conditions are not simply incremental improvements to a common design but are structured to isolate specific theoretical mechanisms \citep{carless2018development, chong2021reconsidering, er2021collaborative, gladovic2025feedback}.

In the first condition, \textit{Directed Feedback}, students received static AI-generated feedback comments without structured support for their use, establishing a baseline for the presentation of comments alone \citep{carless2018development, chong2021reconsidering, ziqi2024l2, er2021collaborative, wood2023enabling}.

In the second condition, \textit{Self-Directed Feedback}, students had access to optional AI assistance, but all interaction remained entirely student-initiated, isolating whether the availability of AI-supported dialogue, absent structural scaffolding, is sufficient to shift engagement \citep{wood2023enabling, brummernhenrich2025applying, usher2025generative}.

In the third condition, \textit{Enacted Feedback}, AI-generated feedback comments were embedded within a staged workflow that prompted students to select suggestions, evaluate their relevance and applicability, and engage in targeted AI-supported dialogue anchored to their selections \citep{carless2018development, chong2021reconsidering, er2021collaborative, wood2023enabling, gladovic2025feedback, usher2025generative, liu2025ai}.

This condition operationalises the theoretical argument that productive feedback encounters must be intentionally structured rather than assumed \citep{winstone2019designing, er2021collaborative}, and that behaviours associated with feedback literacy can be scaffolded through structured participation in feedback processes rather than treated solely as pre-existing attributes of the student \citep{carless2018development, chong2021reconsidering, malecka2022eliciting, wood2023enabling, gladovic2025feedback}.

The three conditions examined in the present study provide an empirical basis for testing
whether the educational value of AI-generated feedback comments depends primarily on
the provision of comments, the availability of AI-supported dialogue, or structured support for feedback use. The present study therefore evaluates the three
workflows across 13{,}037 students in authentic higher education settings,
focusing on students' behavioural engagement with the workflows, self-assessment confidence, and
submitted-work quality.

\section{Literature review}
\subsection{The feedback problem: Provision, use, and the gap between them}
Feedback is widely recognised as an important influence on student learning and achievement \citep{hattie2007power, nicol2021power}. When feedback processes function effectively, they help learners compare their current performance with desired standards and support evaluative and revision activity that can drive improvement \citep{nicol2021power, carless2018development}. However, realising this potential in education involves two distinct challenges: generating high-quality feedback comments at scale and supporting students to use those comments productively.

The provision challenge is structural. Generating timely, individualised, and actionable feedback comments requires subject-matter knowledge, familiarity with students' work, and considerable educator time \citep{boud2013rethinking}. In large-enrolment courses, growing class sizes and workload pressures can limit the timeliness, specificity, and usefulness of comments for subsequent work \citep{winstone2019designing}. Students may consequently encounter grades with limited commentary, comments that arrive too late to inform revision, or suggestions that are difficult to translate into action \citep{carless2018development}. These limitations reflect longstanding resource constraints rather than simply shortcomings in individual educators' practices.

The use challenge persists even when comments are timely, specific, and well crafted. Students may not understand comments, may disengage from those they find confusing or discouraging, or may show little evidence of using them in subsequent work \citep{winstone2017supporting, carless2018development}. Productive engagement requires students to interpret evaluative information, compare it with their own understanding, judge which suggestions are relevant, manage affective responses, and translate those judgements into revision decisions \citep{nicol2021power, carless2018development, tai2018developing}. These processes are cognitively and affectively demanding, and their enactment also depends on whether the feedback environment creates meaningful opportunities to act \citep{carless2022teacher, winstone2019designing}. Making high-quality comments available is therefore necessary but does not by itself produce productive feedback use.

Feedback literacy provides one explanation for this gap. It refers to the understandings, capacities, and dispositions that enable students to make productive use of feedback information \citep{carless2018development}. \citet{carless2018development} identify four interrelated dimensions: appreciating feedback, making judgements, managing affect, and taking action. These capabilities intersect with self-regulated learning (SRL), through which students monitor their performance, regulate cognitive, motivational, and affective processes, and adapt their strategies in pursuit of learning goals \citep{zimmerman2002becoming, panadero2017review}. Within feedback encounters, interpreting comments, judging their relevance, managing affective responses, and deciding how to revise can therefore form part of the regulatory processes connecting feedback information with subsequent action \citep{nicol2006formative}. Empirical evidence associates students' feedback literacy with the uptake of tutor comments and improvements in assessment performance \citep{karunarathne2024evaluating}. More recent scholarship treats feedback literacy as developmental and shaped by the learning environment rather than as a fixed individual attribute \citep{pitt2023enabling, chong2021reconsidering, winstone2022discipline}. For example, \citet{wu2026different} identify dialogue, teacher instruction, and prior knowledge as factors shaping students' developmental trajectories through feedback processes involving elicitation, processing, and enactment. Feedback environments therefore need to create structured opportunities for students to develop and exercise these capabilities.

The distinction between provision and use is particularly consequential for research on AI-generated feedback comments. GenAI can relax constraints on the production of timely and individualised comments, but this capability does not ensure their uptake \citep{pozdniakov2026, bearman2024developing, yan2025distinguishing}. The educational problem therefore concerns not only whether GenAI can generate pedagogically sound comments but also how AI-mediated feedback workflows support students to engage with them.

\subsection{AI-generated feedback: Promise and persistent limitations}
Reviews and empirical studies indicate that GenAI can produce individualised, task-specific, and immediately available feedback comments that can meet or approach the pedagogical quality of educator-generated comments in a range of contexts \citep{banihashem2024feedback, pozdniakov2026, er2025assessing}. A systematic review by \citet{moore2024harnessing} found that GenAI-based automated feedback can support diverse instructional purposes, reduce educator workload, enhance communication, and offer cognitive and emotional support. \citet{chan2024generative} also reported improvements in university students' essay quality, engagement, and motivation during revision following AI-generated feedback comments.

Several affordances underpin this potential. GenAI systems can make comments available during drafting, when students still have an immediate opportunity to revise \citep{moore2024harnessing}. When guided by explicit prompts and assessment criteria, they can generate structured, criteria-based comments with less variability across submissions \citep{kondo2026ai}. Their scalability also allows individualised comments to be generated for large cohorts without a corresponding increase in educator workload \citep{banihashem2024feedback, moore2024harnessing}. Together, these affordances can substantially reduce the provision constraint, particularly in large-enrolment settings.

Students may value the specificity and clarity of AI-generated feedback comments, but they also report concerns about accuracy, source credibility, and over-reliance \citep{guardia2026human,nazaretsky2026gives}. They may dismiss useful comments because they distrust the source or, conversely, accept suggestions uncritically without exercising their own judgement \citep{guardia2026human, nazaretsky2026gives}. GenAI can also produce comments that are fluent and plausible yet imprecise, contextually inappropriate, or incorrect \citep{bearman2024developing}. Students must therefore be able to assess their accuracy and relevance rather than equate fluency with quality.

Evidence of limited uptake reinforces this concern. Several studies report that students do not consistently engage with or act on AI-generated feedback comments, including those judged to be of reasonable quality \citep{pozdniakov2026, guo2025peer, zou2026investigating, tam2025interacting}. \citet{yan2025distinguishing} similarly argues that interaction quality, rather than access alone, shapes whether GenAI supports learning. Strategic prompting, iterative interaction, and verification have been associated with more productive use, whereas unguided access may produce surface-level gains without deeper learning \citep{cong2025critical, qian2025pedagogical, hon2026generative}. Conversational access can allow students to seek clarification, ask follow-up questions, and explore possible revisions, and deeper interaction with GenAI-based chatbots and adaptive systems has been associated with learning, self-efficacy, and engagement \citep{Liang2023The, Bai2025Impact}. However, students' use of these affordances varies with context, prior knowledge, motivation, attitudes, and perceived usefulness \citep{Li2025The, Almogren2024Exploring, Dahri2024Investigating, khosravi2026building}. Open-ended dialogue may therefore benefit students who already know what to ask and how to evaluate the response more than those who need greater support \citep{qian2025pedagogical}.

\subsection{Designing feedback enactment: Agency, dialogue, and evaluative judgement}
An ecological view of feedback literacy locates the development of feedback-literate behaviour in students' participation in purposefully structured feedback encounters \citep{pitt2023enabling, chong2021reconsidering, winstone2022discipline}. Whether students recognise a need for support, seek feedback, evaluate the information received, and translate it into revision also involves SRL \citep{zimmerman2002becoming, panadero2017review}. Within feedback encounters, the capacities described by feedback literacy are exercised through these regulatory processes: evaluative judgement supports assessment of the quality and relevance of feedback information, while subsequent action requires students to decide how that information should inform their work \citep{nicol2006formative, tai2018developing}. The workflow surrounding AI-generated feedback comments can therefore shape feedback enactment by supporting regulatory processes that students would otherwise need to initiate and manage independently.

Student agency concerns students' meaningful control over their engagement with feedback information. Students exercise agency when they make consequential decisions about which suggestions to prioritise, which aspects of their work need attention, and which comments are relevant to their circumstances \citep{nicol2021power, nicol2024shifting}. Such decisions can activate internal comparison processes through which students compare external information with their own understanding and generate internal feedback \citep{nicol2021power}. Embedding these decisions in a workflow allows students to exercise agency while receiving structural support.

Dialogue enables students to clarify comments, test their understanding, explore alternatives, and negotiate meaning rather than encounter feedback as a one-way transmission \citep{ajjawi2018examining, nedrehagen2025scoping, myers2025dialogism}. \citet{myers2025dialogism} distinguish among clarificatory dialogue, which addresses what comments mean; questioning dialogue, which examines why a suggestion matters; and critical dialogue, which evaluates whether and how it should be applied. Conversational agents can support iterative questioning and explanation in AI-mediated environments \citep{zhai2023systematic}. However, open-ended access leaves students to identify which points warrant discussion and formulate productive questions. Anchoring dialogue to comments that students have selected can connect conversational interaction to a specific evaluative and revision purpose.

Evaluative judgement is the capacity to assess the quality of one's own and others' work \citep{tai2018developing, ilangakoon2022relationship}. In this context, that capacity was exercised through judgements about the relevance and applicability of AI-generated suggestions. It is both required for productive feedback use and developed through practices such as dialogue and self-assessment \citep{ilangakoon2022relationship}. Prompting students to judge which suggestions matter and explain why before revision can encourage evaluation rather than uncritical acceptance and reduce the outsourcing of judgement to the AI system \citep{bearman2024developing}.

Taken together, agency, dialogue, and evaluative judgement provide mechanisms through which AI-mediated feedback workflows can support students' self-regulation. Such workflows can create structured opportunities for students to select and prioritise comments, evaluate their relevance, engage in dialogue anchored to specific points, and translate their judgements into revision. Static comments do not require students to engage in these processes, while optional conversational access makes them possible but leaves their enactment to student initiative \citep{carless2018development, winstone2019designing, bearman2024developing, moore2024harnessing}. A structured workflow can instead scaffold the monitoring, decision-making, and action involved in feedback enactment.

Related empirical and conceptual work supports this principle. \citet{Quinton04072025} found that guidance within peer-feedback processes produced commentary perceived as more useful and trustworthy than unstructured, open-ended tools. \citet{yang2025feedback} found that feedback literacy predicted both the uptake and perceived usefulness of AI-generated feedback comments. \citet{nicol2024shifting} further argues that feedback agency develops when students have structured opportunities to compare their work with reference information and generate self-feedback.

Despite growing conceptual and empirical interest in AI-mediated feedback, direct comparative evidence on how the structure of AI-mediated feedback workflows shapes students' enactment of feedback in authentic higher education contexts remains limited. In particular, it is not yet clear how variations in workflow structure are associated with behavioural engagement with feedback workflows, self-assessment confidence, and submitted-work quality. By comparing three AI-mediated feedback workflows at scale, the present study addresses this gap and provides a direct empirical test of the proposition that productive AI-mediated feedback encounters require intentional workflow structure rather than merely making feedback comments or AI assistance available.

\section{Method}

This study examined how the structure of AI-mediated feedback 
workflows was associated with students' behavioural engagement, self-assessment
confidence, and submitted-work quality. Three
theoretically distinct workflows were implemented within the same platform for
student-authored learning resources: 
\begin{itemize}
    \item \textit{Directed Feedback}, which operationalises
the presentation of static AI-generated feedback comments without structured support for their use;
\item \textit{Self-Directed
Feedback}, which isolates the availability of optional, student-initiated AI assistance, testing whether
access to AI-supported dialogue, absent structural scaffolding, is sufficient
to support students' engagement;
\item  \textit{Enacted Feedback}, which
operationalises structured support for feedback enactment by embedding opportunities for student agency, evaluative
judgement, and selection-anchored AI-supported dialogue directly into the workflow.
\end{itemize}

The comparison is guided by the following research questions:
\begin{itemize}
    \item \textbf{RQ1: Uptake, revision counts, and event-flow transitions.} How do the three AI-mediated feedback
    workflows compare in terms of workflow-specific uptake, revision counts, and
    event-flow transitions?
    \item \textbf{RQ2: Self-assessment confidence.} How do the three AI-mediated feedback
    workflows compare in terms of students' self-assessment confidence?
    \item \textbf{RQ3: Submitted-work quality.} How do the three AI-mediated feedback
    workflows compare in terms of students' submitted-work quality?
\end{itemize}
 
This study was conducted under institutional ethics approval from The
University of Queensland [2023/HE001453]. Participation in RiPPLE was a component
of regular coursework, and all data were collected as part of normal platform
operation. Log data were de-identified prior to analysis, and no additional
data collection instruments were administered for the purposes of this study.
The four subsections that follow describe the
research tool, the three AI-mediated feedback workflows, the study design and
dataset, and the approach to data analysis.

\subsection{The RiPPLE platform}

RiPPLE (Recommendation in Personalised Peer-Learning Environments) is an
adaptive learning platform that engages students in creating, evaluating,
revising, and practising with peer-generated learning resources
\citep{khosravi2019ripple}. In RiPPLE, students both contribute educational
resources that support their peers' learning and benefit from the cognitive
processes involved in authoring, reviewing, and refining
learning materials. This creates a learning environment in which students are
not only consumers of instructional content but also active contributors to
the shared pool of resources used within the course.

RiPPLE structures student activity across three connected stages:
\textit{creation}, \textit{review}, and \textit{practice}. Each stage serves a
distinct pedagogical function, and the platform's analytics layer connects
these functions so that resources produced during creation can be moderated,
refined, and reused in subsequent learning activities. In the creation stage,
students author learning resources. In the review stage, students evaluate
resources created by their peers. In the practice stage, approved resources are
used by other students for revision and learning practice. RiPPLE has been
deployed in higher education courses across multiple disciplines, and prior
research has examined its effects on learning gains, engagement, and the
quality of student-authored learning resources \citep{khosravi2019ripple}.

\begin{figure}
    \centering
    \includegraphics[width=0.75\linewidth]{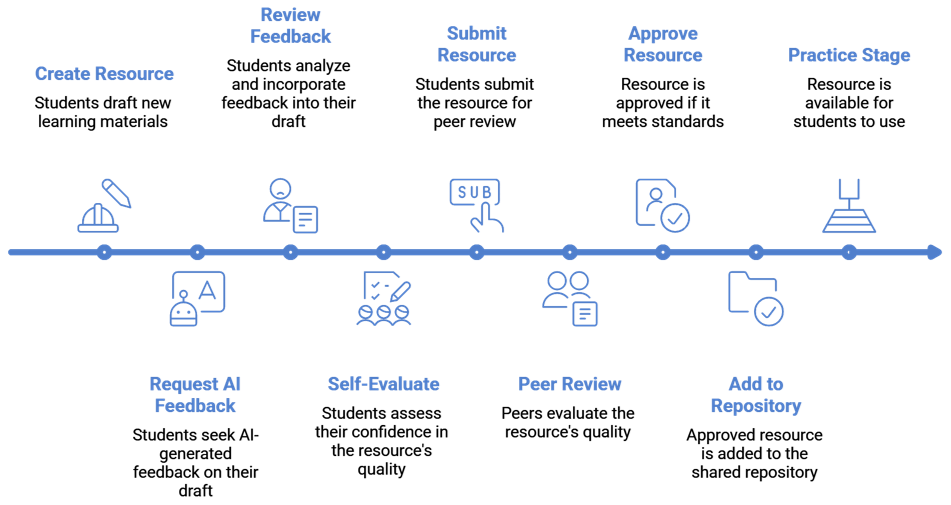}
    \caption{The general RiPPLE resource creation sequence in conditions that provided AI-generated feedback comments: students draft a resource, access and review the comments, revise the draft, complete self-assessment, and submit the resource for peer moderation.}
    \label{fig:ripple}
\end{figure}

In the \textit{creation} stage, students develop learning resources such
as multiple-choice questions, worked examples, flashcards, notes, and
short slide-based explanations. Authoring is supported through a
structured interface that prompts students to specify the topic,
difficulty level, and learning objective associated with each resource.
During this stage, the platform can provide AI-mediated support before students
finalise the resource. Depending on the condition, this support took the form of
AI-generated feedback comments or optional AI assistance. Where generated, the
feedback comments were produced by a large language model conditioned on the
resource type, topic metadata, and authoring guidelines and were presented as
formative commentary organised around strengths and suggestions for improvement.
The specific model version differed across implementation periods to reflect the
technology available at each deployment, as detailed in
Section~\ref{sec:workflows}. The structure of the AI-mediated feedback workflow
is the focal intervention in the present study, and the three workflows described
in Section~\ref{sec:workflows} differ in how AI-generated feedback comments or
optional AI assistance are encountered and used. Figure~\ref{fig:ripple} shows
the general resource creation sequence for conditions that provided AI-generated
feedback comments.

In the \textit{review} stage, submitted resources undergo peer
moderation. Each resource is evaluated by multiple student moderators
who rate its quality against a structured rubric covering accuracy,
clarity, alignment with the stated topic, and pedagogical value.
Peer moderation ratings are aggregated into a single submitted-work quality
score on a 0--5 scale, which determines whether a resource is approved for
inclusion in the shared course repository. Peer moderation serves two
roles in the platform: it acts as a quality-control mechanism for
student-authored learning resources, and it provides an authentic assessment context
in which the educational value of students' work can be observed. In the present study, these peer moderation outcome scores are used as the
operational measure of submitted-work quality (RQ3).

In the \textit{practice} stage, approved resources are surfaced to other
students through an adaptive recommendation engine that selects items
based on each student's current mastery of the relevant topics. This
stage closes the loop between authoring and learning: resources created
and moderated by students become part of the practice material used by
their peers. Although the practice stage is not the focus of the present
analysis, it is relevant because it establishes that the quality of
authored resources has downstream consequences for the wider learning
community, thereby providing authentic motivation for students to engage
with feedback comments during the creation stage.

RiPPLE was a suitable platform for the present study for three reasons.
First, its architecture made it possible to vary the AI-mediated feedback
workflow at the platform level while holding the authoring
task, peer moderation procedure, and quality rubric constant across
conditions. This was essential for isolating the effect of workflow
structure from the effect of comment content or task design.
Second, the platform's instrumentation captured fine-grained interaction
logs that record every state transition between drafting, reviewing feedback
comments, AI-supported dialogue, revision, and submission. These logs enabled the measures of workflow-specific uptake, revision counts, event-flow transitions, self-assessment confidence, and submitted-work quality used to address all research questions. Third, the platform's authentic use in credit-bearing courses meant that students were engaging with the feedback process in a setting where their submitted work had real consequences, supporting the ecological validity of the comparisons across the three conditions.

\subsection{AI-mediated feedback workflows}
\label{sec:workflows}
This study compared three AI-mediated feedback workflows implemented within RiPPLE. The comparison focused on how AI-generated feedback comments or optional AI assistance were integrated before final submission. Across all three conditions, students created a draft learning resource, had access to an AI-mediated support process, completed self-assessment, and submitted the resource for moderation. The key distinction between conditions was the degree to which the platform structured students' engagement after the initial draft.

\begin{figure}[htbp]
    \centering
    \includegraphics[width=0.95\linewidth]{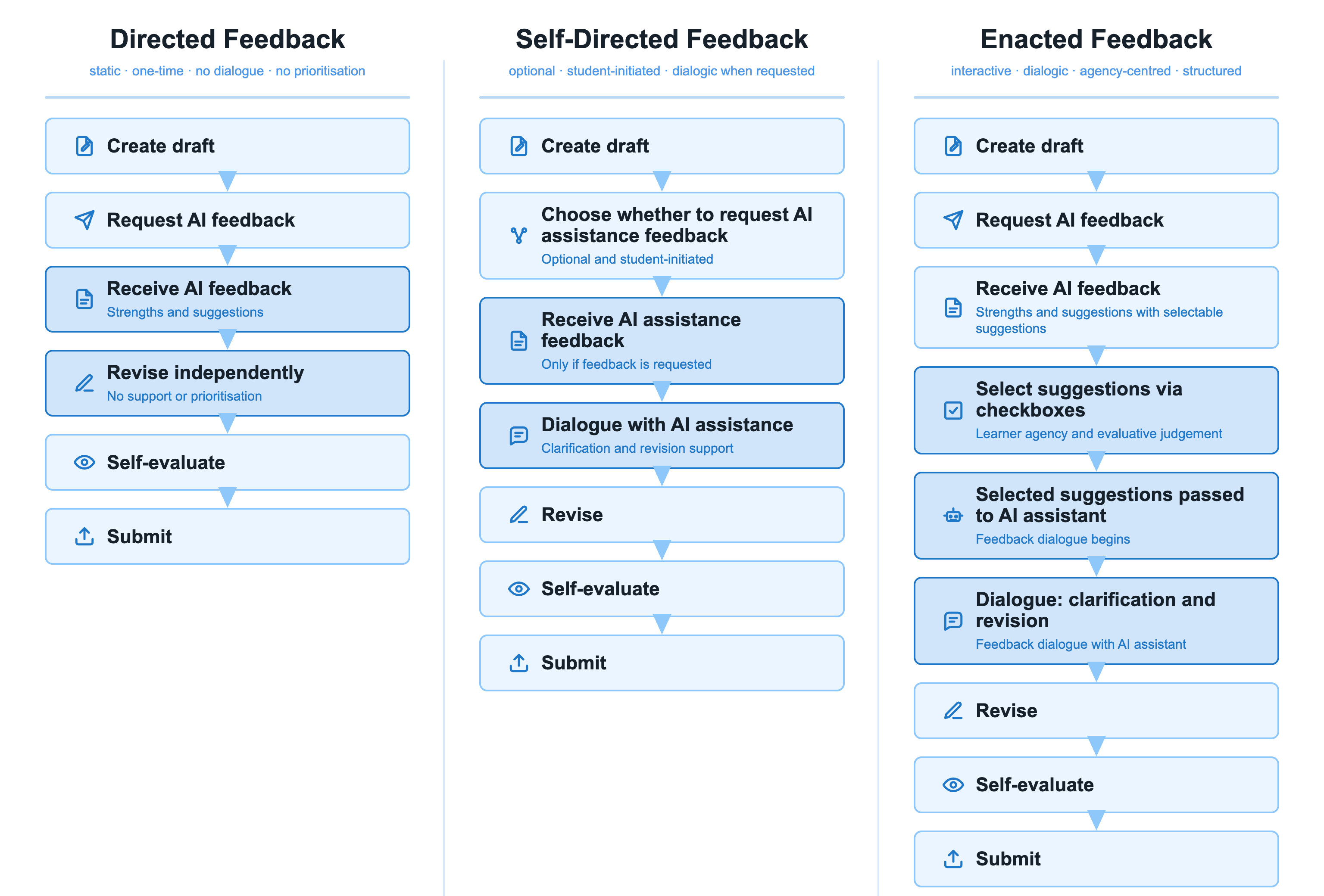}
    \caption{Comparison of the three AI-mediated feedback workflows: \textit{Directed Feedback}, \textit{Self-Directed Feedback}, and \textit{Enacted Feedback}.}
    \label{fig:compare}
\end{figure}

The three conditions represented distinct forms of workflow support. The \textit{Directed Feedback} condition provided a one-time set of static AI-generated feedback comments without an AI-supported dialogue pathway. The \textit{Self-Directed Feedback} condition introduced optional access to AI-supported dialogue, but interaction remained entirely student-initiated and was not organised around selected suggestions. The \textit{Enacted Feedback} condition embedded AI-generated feedback comments within a staged process that prompted students to select suggestions and use those selections as the basis for targeted AI-supported dialogue. In this way, the three conditions operationalised a progression from static comments to optional AI assistance and then to comments embedded in a scaffolded feedback-use process. Figure~\ref{fig:compare} summarises the three workflows.

The underlying language model used for AI-mediated support differed across implementation periods, reflecting the model versions available at the time of each deployment. The \textit{Directed Feedback} condition (Semester 1, 2025) and the \textit{Self-Directed Feedback} condition (Semester 2, 2025) used OpenAI's GPT-4o mini. The \textit{Enacted Feedback} condition (Semester 1, 2026) used GPT-5 mini. In the two conditions that presented AI-generated feedback comments, the prompt structure, conditioning metadata, and output format (strengths and suggestions for improvement) were held constant.

\subsubsection{\textit{Directed Feedback} workflow}

\begin{figure}[htbp]
    \centering
    \includegraphics[width=0.60\linewidth]{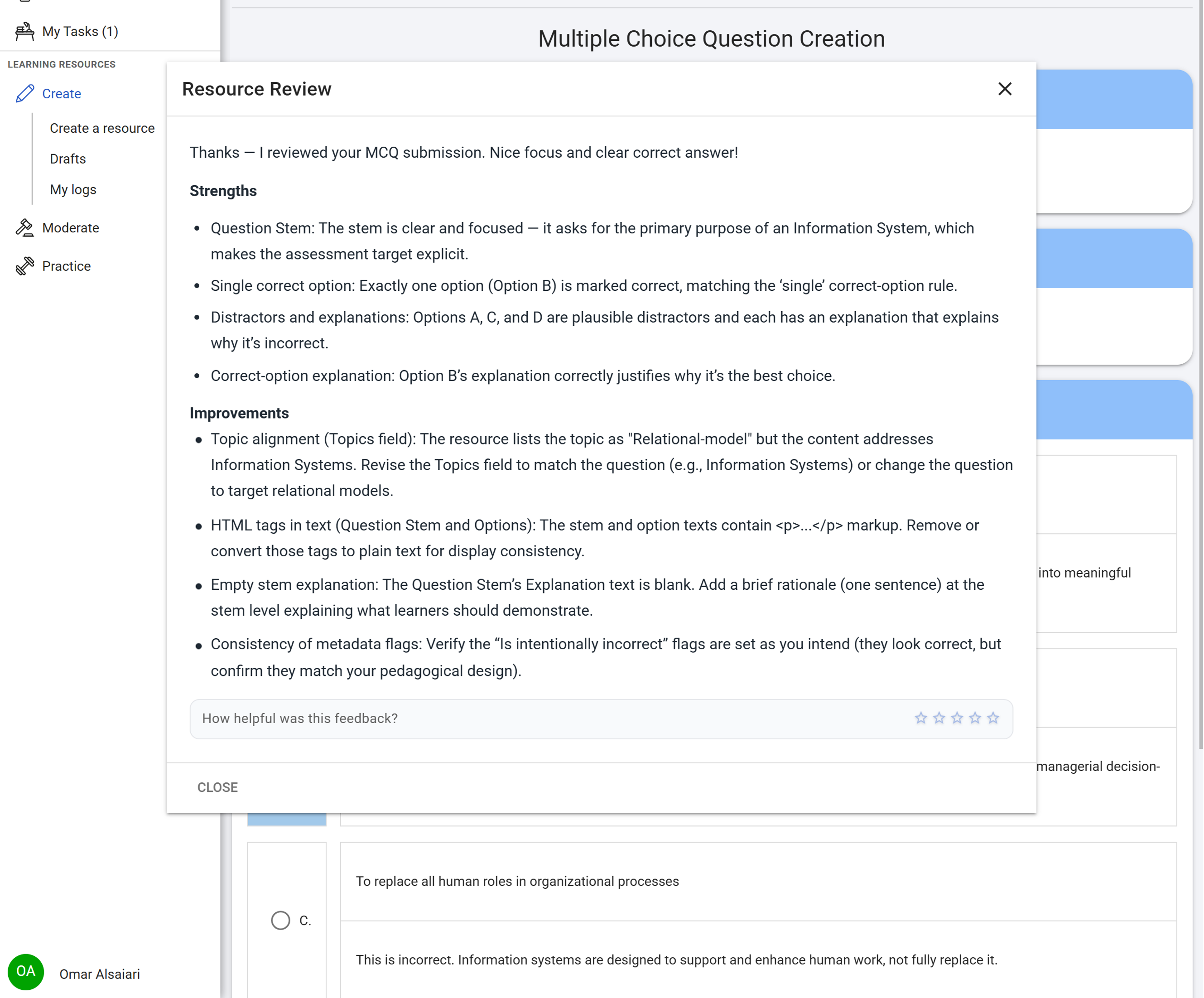}
    \caption{\textit{Directed Feedback} interface: one-time static AI-generated feedback comments without dialogue or structured prioritisation.}
    \label{fig:directed_feedback}
\end{figure}

The \textit{Directed Feedback} workflow operationalised the presentation of static AI-generated feedback comments without structured support for their use. It established a baseline condition in which comments were presented to students without an AI-supported dialogue pathway. In this condition, students requested AI-generated feedback comments on a draft learning resource and received a single static review within the RiPPLE authoring interface, as illustrated in Figure~\ref{fig:directed_feedback}. The comments were organised into two sections, strengths and suggestions for improvement. Their function was formative: to provide guidance on the gap between the current draft and acceptable standards before final submission, without constituting a requirement to revise.

The interface presented the AI-generated feedback comments as a read-only document. After reading the comments, students could proceed directly to self-assessment and submission, or they could return to the editor and revise their draft before completing the self-assessment step. No mechanism required students to engage with specific suggestions, indicate which suggestions they intended to act on, or explain the rationale for any revision choices they made. The interface also provided no dialogic pathway: students could not ask follow-up questions, seek clarification
on a suggestion, or request elaboration on any aspect of the review. The comments were generated once, at the point of request, and remained visible in the interface during the subsequent editing session, but no further AI interaction was possible within this workflow.

\subsubsection{\textit{Self-Directed Feedback} workflow}

\begin{figure}[htbp]
    \centering
    \includegraphics[width=0.70\linewidth]{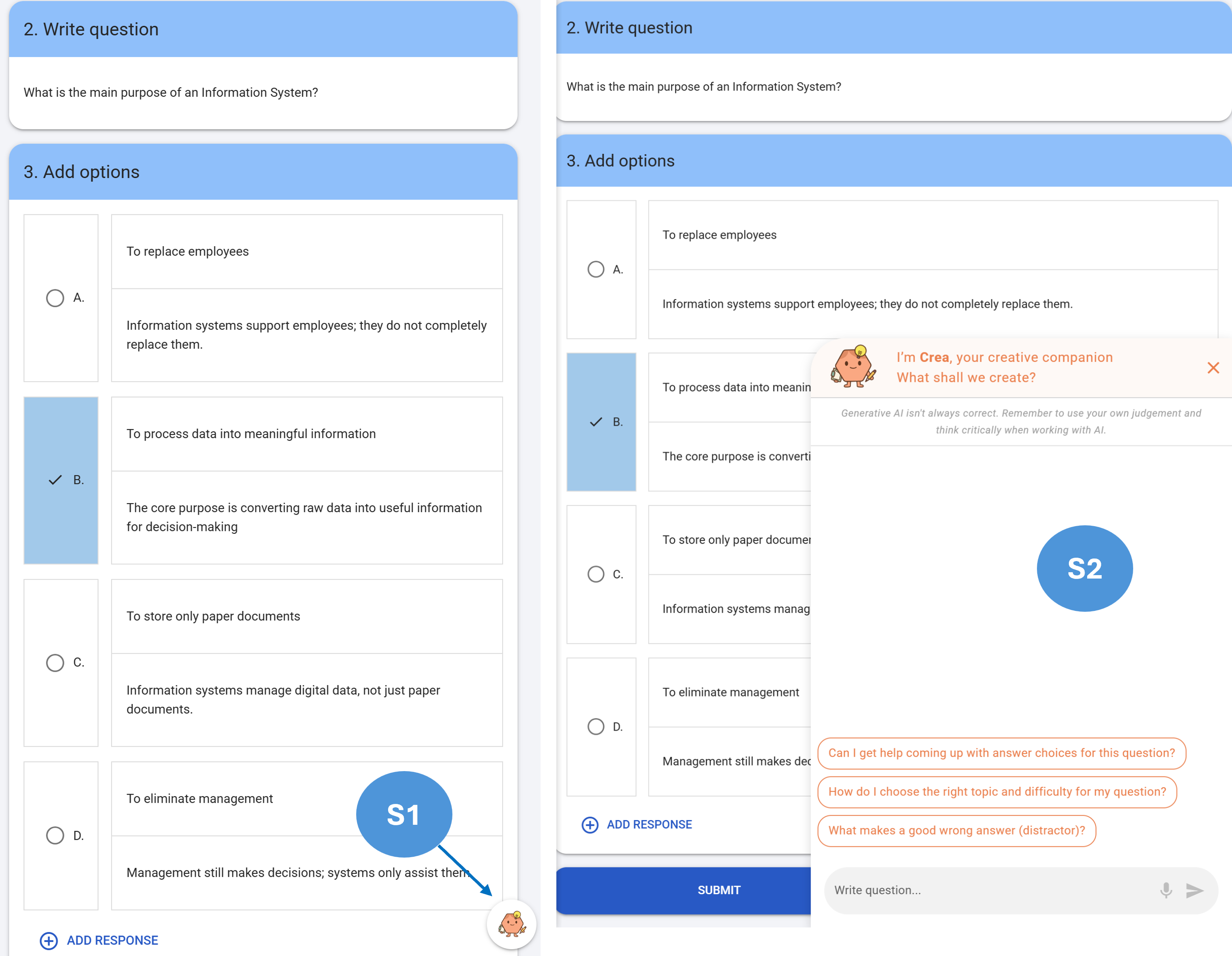}
    \caption{\textit{Self-Directed Feedback} interface: an optional AI assistance workflow in which students initiate dialogue for clarification and revision support before submission.}
    \label{fig:self_directed_feedback}
\end{figure}

The \textit{Self-Directed Feedback} workflow was designed to isolate the availability of optional AI-supported
dialogue and test whether access to optional AI assistance, absent
structural scaffolding, was sufficient to shift engagement. The workflow introduced an AI-supported dialogue
pathway within the resource creation process. The workflow consisted of
two states: a standard drafting interface in which AI assistance was
available but student-initiated (S1), and an active AI-supported dialogue panel
that opened when the student chose to engage (S2).
Figure~\ref{fig:self_directed_feedback} illustrates both states side
by side.

In S1, students drafted their learning resource within the standard
authoring interface. An AI assistance icon was visible within the
interface, signalling that AI-supported dialogue was available, but no
feedback comments were generated automatically and no prompt was issued to
initiate dialogue. Whether to open the AI assistance panel, when to do
so, and what to ask were left entirely to the student's discretion.
Unlike the \textit{Directed Feedback} condition, in which students could request a static review of their draft, this workflow generated no AI-generated feedback comments. Instead, students could access AI-supported dialogue at
any point during drafting. However, because the system provided no
structured guidance on how or when to engage, the relevance and quality
of any interaction depended on each student's existing capacity to
self-diagnose what they needed.

In S2, students who chose to engage were presented with the AI
assistance dialogue panel. The panel offered a set of
suggested prompts, such as requests for help generating answer choices,
guidance on selecting an appropriate topic or difficulty level, and
criteria for constructing plausible distractors, as well as a free-text
input field for student-authored queries.

This workflow differed from the \textit{Directed Feedback} workflow in that
students were not limited to a single static review; they could initiate
repeated and open-ended interaction with the AI assistance panel across
the drafting process. It differed from the \textit{Enacted Feedback} workflow in
that interaction was not anchored to specific, selected suggestions. The system made dialogue available but did not structure what
students asked, which suggestions they should prioritise, or how they
should translate AI responses into revision. This condition was therefore
analytically important because it isolated the availability of AI-supported
dialogue from structured support for feedback enactment, allowing
the study to determine whether access to AI-supported dialogue, absent
scaffolding, is sufficient to shift engagement and improve outcomes.

\subsubsection{\textit{Enacted Feedback} workflow}

\begin{figure}[htbp]
    \centering
    \includegraphics[width=0.90\linewidth]{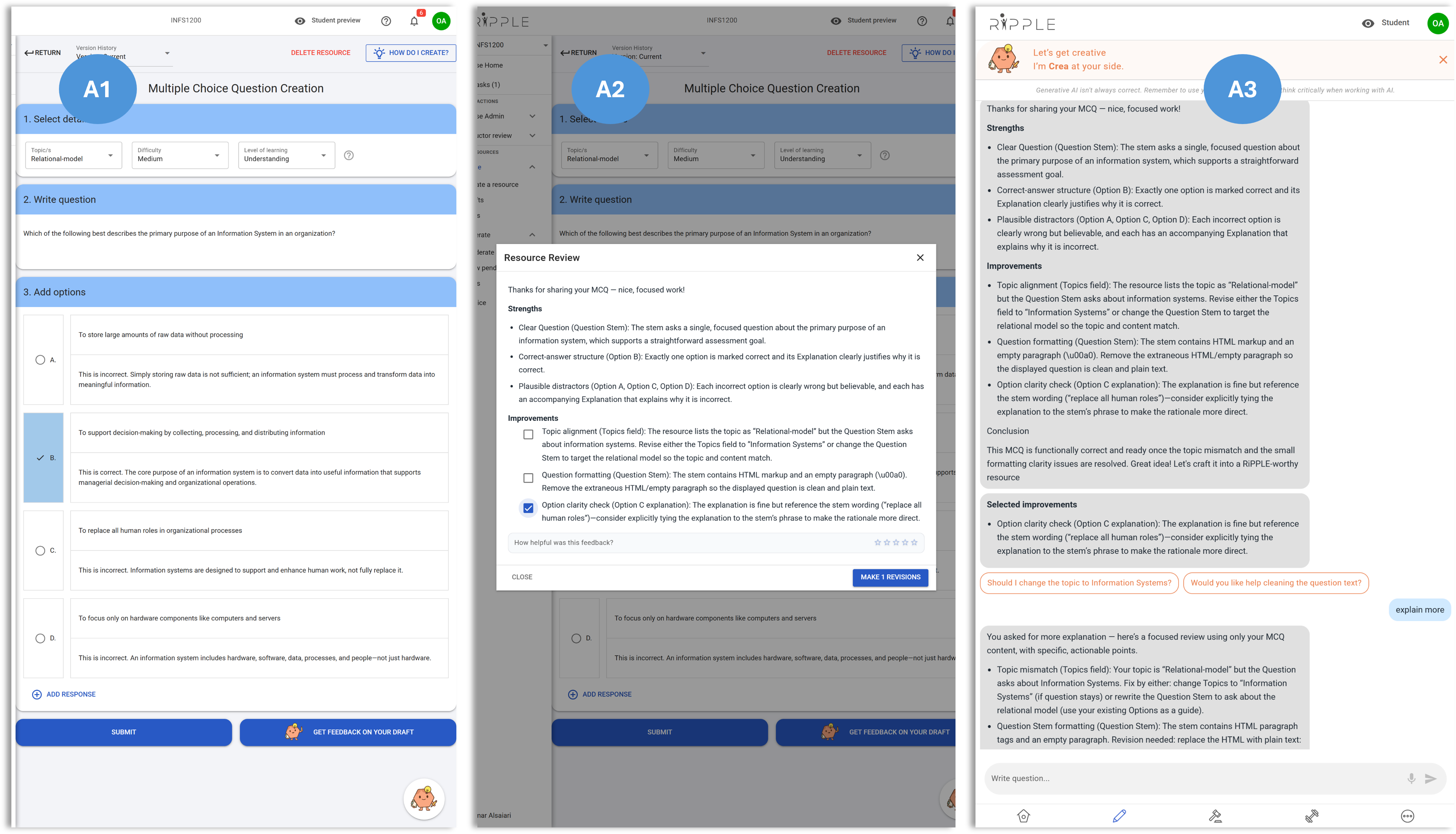}
    \caption{\textit{Enacted Feedback} interface: an AI-mediated feedback workflow in which students receive initial feedback comments and are prompted to select improvement suggestions and engage in targeted AI-supported dialogue.}
    \label{fig:enacted_feedback}
\end{figure}

The \textit{Enacted Feedback} workflow operationalised structured support for feedback enactment,
embedding explicitly prompted opportunities for student agency, evaluative judgement, and selection-anchored AI-supported dialogue. The \textit{Enacted Feedback} workflow was designed on the premise that productive engagement with AI-generated feedback comments cannot be assumed to follow from the provision of comments alone, but must be structured into the
learning environment through deliberate design
\citep{winstone2019designing, carless2018development}. Rather than
treating comments as a completed output to be read and optionally acted
upon, this workflow positioned feedback comments as the entry point to a staged
process that prompted students to select, evaluate, discuss,
and revise before submission while retaining the option to bypass the structured pathway. The workflow scaffolded three
behaviours associated with productive feedback use that the \textit{Directed Feedback} and \textit{Self-Directed Feedback}
conditions left to students' discretion: student agency through
selection, evaluative judgement through prioritisation, and dialogic
engagement through targeted AI assistance
\citep{carless2018development, tai2018developing, myers2025dialogism}.
As shown in Figure~\ref{fig:enacted_feedback}, the interaction was structured
across three sequential stages: receiving initial AI-generated feedback comments
(A1), selecting improvement suggestions to address (A2), and entering a
focused AI-supported dialogue anchored to the selected suggestions (A3).

In A1, students received AI-generated feedback comments within the draft
authoring interface. The comments retained the same general structure as
the \textit{Directed Feedback} condition, comprising strengths and suggestions
for improvement, organised around the same resource-type and topic
metadata. The A1 stage served both as the
informational input to the subsequent stages of the \textit{Enacted Feedback}
workflow and as the comparability anchor for the structural format of
the initial comments.

In A2, the suggestions for improvement presented in A1 were transformed
from a read-only list into a set of selectable items displayed alongside
the draft resource. At this stage, students faced a branching decision
with three available paths. They could select one or more suggestions
and proceed to the targeted AI-supported dialogue interface (A3), where
interaction would be anchored to their selected suggestions. They
could close the feedback panel without selecting, returning to the draft
editor. Alternatively, they could navigate directly to the self-assessment step,
bypassing further engagement with the feedback panel entirely. In all
three paths, students retained the ability to revise their draft and to
initiate AI assistance before submission; what differed was whether that
AI assistance was anchored to selected suggestions or remained
unstructured and self-directed. Students who closed the panel or
proceeded to self-assessment without selecting could still edit their
resource and open the AI assistance interface, but the dialogue in those
cases was not contextualised by prior selection and therefore resembled
the optional, student-initiated interaction available in the
\textit{Self-Directed Feedback} condition rather than the structured enactment
pathway of A3.

For students who engaged with it, the selection step was designed to activate two specific
behaviours associated with feedback literacy. First, it supported
student agency by giving students meaningful control over which
suggestions they chose to pursue, rather than presenting an undifferentiated
set of comments to interpret independently
\citep{nicol2021power, nicol2024shifting}. Second, it supported
evaluative judgement by inviting students to make an explicit decision
about the value, relevance, and applicability of each suggestion to
their specific draft before committing to further interaction
\citep{tai2018developing, bearman2024developing}. In learning
environments where AI-generated feedback comments may appear authoritative even when they
require critical scrutiny, this requirement to judge before acting
provided a structural prompt for the kind of reflective evaluation that
feedback literacy research identifies as a prerequisite for productive
feedback use \citep{bearman2024developing, yang2025feedback}. The
non-selection paths, meanwhile, preserved the ecological validity of
the workflow: students who closed the panel or proceeded directly to self-assessment were
not forced into dialogue, and their subsequent AI interactions, if any,
remained unanchored, allowing the study to observe naturally occurring
variation in how students chose to engage with the available support
structures.

In A3, the suggestions selected in A2 were automatically passed into
the AI assistance dialogue interface, which opened as an embedded panel
alongside the draft editor. The dialogue was
anchored to the specific suggestions the student had selected,
meaning that the AI assistant was provided with context about the student's
chosen priorities rather than engaging in a general, open-ended
conversation about the resource. Within
this anchored dialogue, students could ask for clarification of the
meaning of a suggestion, request elaboration on why a particular element
of their draft required revision, explore alternative ways to address a
suggestion, and develop concrete strategies for translating the
suggestion into specific changes to the draft. This form of
selection-anchored dialogue addresses a limitation identified in
research on unstructured AI-supported dialogue: students who already know what to
ask and how to interpret responses benefit disproportionately from
open-ended conversational access, while students with less developed
feedback literacy may use the dialogue superficially or not at all
\citep{guo2025peer, yan2025distinguishing}. By anchoring the
dialogue to selected suggestions, the A3 interface aimed to reduce the
burden of self-directing the conversation and channelled interaction
toward the specific revision decisions students had already indicated
they intended to make. Following the AI-supported dialogue, students could revise
their draft, complete self-assessment, and submit the resource for
moderation, consistent with the shared pipeline across all three
conditions.

\subsection{Study design and dataset}

The study used a quasi-experimental sequential
cohort design in which each implementation period corresponded to one
feedback workflow condition. The \textit{Directed Feedback} condition was implemented
in Semester 1, 2025, the \textit{Self-Directed Feedback} condition in Semester 2,
2025, and the \textit{Enacted Feedback} condition in Semester 1, 2026. This design
was appropriate because the intervention was implemented at the level of the
platform workflow rather than randomly assigned to individual students within
the same course offering. To support comparability across conditions, the
same platform, task structure, moderation rubric, and
assessment logic were retained throughout the implementation periods. In all
three conditions, students completed the same general resource creation
process: they authored learning resources, engaged with 
the condition-specific feedback or assistance workflow before submission, completed
self-assessment, and submitted resources for moderation. The primary design
difference was therefore the nature of the AI-mediated feedback workflow
available before submission.

The dataset comprised 13{,}037 students, 51{,}296
student-authored resources, and 70 course offerings across the three feedback
workflow conditions. Specifically, the \textit{Directed Feedback} condition included
3{,}723 students, 14{,}425 resources, and 21 course offerings; the
\textit{Self-Directed Feedback} condition included 3{,}951 students, 15{,}548 resources,
and 25 course offerings; and the \textit{Enacted Feedback} condition
included 5{,}363 students, 21{,}323 resources, and 24 course
offerings. Table~\ref{tab:condition_descriptives} reports the number of
students, resources, and course offerings within each condition.

\begin{table}[htbp]
\centering
\caption{Descriptive summary of students, resources, and course offerings across the three feedback workflow conditions}
\label{tab:condition_descriptives}
\begin{tabular}{lccc}
\toprule
\textbf{Condition} &
\textbf{Students} &
\textbf{Resources} &
\textbf{Course offerings} \\
\midrule
Overall                & 13{,}037 & 51{,}296 & 70 \\
\textit{Directed Feedback}      &  3{,}723 & 14{,}425 & 21 \\
\textit{Self-Directed Feedback} &  3{,}951 & 15{,}548 & 25 \\
\textit{Enacted Feedback}       &  5{,}363 & 21{,}323 & 24 \\
\bottomrule
\end{tabular}
\end{table}

Across all three conditions, students followed the same task structure in
each of the four rounds. In every round, they created one original learning
resource, moderated three peer-created resources, and practised ten
resources authored by other students. This design ensured that students had
comparable exposure to authoring, peer evaluation, and practice activities
across the three conditions.

\subsection{Data analysis}
\label{sec:data_analysis}

All analyses were conducted in R (version 4.4.2). An alpha level of .05 was used throughout. Descriptive statistics were computed before inferential analyses, and outcome distributions were assessed through visual inspection of histograms and model diagnostics. Analyses were conducted at the resource level, with condition entered as a fixed effect. Where students contributed multiple resources, student ID was included as a random intercept to account for repeated observations within students. The omnibus effect of condition in each mixed-effects model was evaluated using a likelihood-ratio test comparing the full model containing condition with an otherwise identical model omitting condition; each model pair was fitted to the same observations and retained the same random-effects structure. For each modelled outcome, pairwise comparisons were obtained from estimated marginal means and reported with Tukey-adjusted simultaneous 95\% confidence intervals and $p$-values. Because the outcomes were binary, count, ordinal, and bounded continuous, effect sizes were reported on the scale native to each fitted model and supplemented by pairwise contrasts on interpretable response scales.

\subsubsection{RQ1: Uptake, revision counts, and event-flow transitions}

Behavioural engagement was assessed through complementary measures: workflow-specific uptake, revision counts, and first-order event-flow transitions. Uptake was operationalised according to the routing structure of each condition. In \textit{Directed Feedback}, uptake was defined as an immediate transition from \texttt{AI Feedback} to an editing state. In \textit{Self-Directed Feedback}, uptake was defined as requesting \texttt{AI Assistance} and subsequently editing the resource, where the edit could occur after one or more intermediate AI assistance states. In \textit{Enacted Feedback}, uptake was defined as any downstream edit after the scaffolded process for AI-generated feedback comments, because the workflow routes students through suggestion-selection and AI-assistance states before revision. A binomial GLMM with a logit link was fitted because uptake was binary. Pairwise effects were reported as odds ratios and as differences in model-estimated uptake probabilities, expressed in percentage points. This measure captures workflow-specific uptake as operationalised by subsequent editing; it does not establish that a specific feedback suggestion was incorporated into the resource.

Revision count was defined as the number of creation-stage edits recorded for each resource after the relevant workflow state. To reduce the influence of extreme log-file artefacts, revision counts were capped at the empirical 99th percentile, corresponding to a maximum of 8 revisions. Revision counts were analysed using a negative-binomial GLMM with a log link because they were sparse, positively skewed, and overdispersed. Pairwise effects were reported as incidence-rate ratios (IRRs; ratios of model-estimated revision counts) and as absolute differences in model-estimated revisions per resource.

Event-flow transitions were analysed using first-order Markov models (FOMMs). For each condition, event sequences began at the relevant feedback or drafting state, and transition probabilities were calculated as the number of observed transitions from one state to the next divided by the total number of outgoing transitions from the source state. Editing states were combined across \texttt{Question}, \texttt{Options}, and \texttt{Question Details}. Because FOMM transitions reflect only immediate next-state moves, the proportion of resources moving directly from an AI assistance state to an editing state may be smaller than the binomial uptake measure, which captures any downstream edit irrespective of intermediate states.

\subsubsection{RQ2: Self-assessment confidence}

Self-assessment confidence was measured using students' ordinal confidence ratings on a 1--5 scale. A cumulative link mixed model (CLMM) with a logit link was fitted to preserve the ordered structure of the scale. Estimated marginal means (EMMs) were reported on the expected rating scale, alongside observed means, medians, and standard deviations. Pairwise effects were reported as cumulative odds ratios and as differences in model-estimated expected ratings on the 1--5 scale.

\subsubsection{RQ3: Submitted-work quality}

Submitted-work quality was operationalised using peer moderation outcome scores, which were analysed as bounded continuous outcomes. Scores were divided by 5 to place them on a 0--1 scale and were constrained to the open interval $(0,1)$ before modelling. A beta GLMM with a logit link was fitted because moderation outcome scores were bounded. EMMs were back-transformed to the original 0--5 scale, and pairwise effects were reported primarily as response-scale EMM differences. Exponentiated logit-scale contrasts were retained as secondary model-native estimates and reported as ratios of expected-proportion odds.

\section{Results}
\subsection{RQ1: Uptake, revision counts, and event-flow transitions}
\paragraph{\textbf{Uptake}.}

\begin{figure}[!htbp]
\centering

\begin{minipage}[t]{0.61\textwidth}
\vspace{0pt}
\centering

\refstepcounter{table}
\label{tab:rq1_uptake}
{\small\textbf{Table~\thetable:} RQ1: Workflow-specific uptake\par}

\vspace{0.35em}

\begin{threeparttable}
\centering
\tiny
\setlength{\tabcolsep}{1.8pt}
\renewcommand{\arraystretch}{1.08}

\textbf{Panel A. Descriptive statistics and model-estimated uptake}

\vspace{0.35em}
\begin{tabular*}{\linewidth}{@{\extracolsep{\fill}}lrrrrrr@{}}
\toprule
\textbf{Condition} &
\textbf{Resources} &
\textbf{Students} &
\textbf{Uptake, $n$} &
\textbf{Uptake, \%} &
\textbf{Est.\ probability} &
\textbf{95\% CI} \\
\midrule
\textit{Directed Feedback}      & 14{,}425 & 3{,}723 & 2{,}716 & 18.8\% & 14.1\% & [13.3\%, 15.0\%] \\
\textit{Self-Directed Feedback} & 15{,}548 & 3{,}951 & 32      & 0.2\%  & 0.1\%  & [0.1\%, 0.2\%] \\
\textit{Enacted Feedback}       & 21{,}323 & 5{,}363 & 6{,}165 & 28.9\% & 26.2\% & [25.3\%, 27.2\%] \\
\bottomrule
\end{tabular*}

\vspace{0.75em}
\textbf{Panel B. Omnibus likelihood-ratio test}

\vspace{0.35em}
\begin{tabular*}{\linewidth}{@{\extracolsep{\fill}}llrrr@{}}
\toprule
\textbf{Outcome} & \textbf{Model} & \textbf{$\chi^2$} & \textbf{df} & \textbf{$p$} \\
\midrule
Uptake & Binomial GLMM & 5115.97 & 2 & $< .001$ \\
\bottomrule
\end{tabular*}

\vspace{0.75em}
\textbf{Panel C. Tukey-adjusted post hoc contrasts}

\vspace{0.35em}
\begin{tabular*}{\linewidth}{@{\extracolsep{\fill}}llllrr@{}}
\toprule
\textbf{Contrast} & \textbf{OR} & \textbf{OR 95\% CI} & \textbf{Difference, pp (95\% CI)} & \textbf{$z$} & \textbf{$p_{\mathrm{adj}}$} \\
\midrule
Enacted / Directed       & 2.16   & [1.96, 2.38]     & +12.1 [10.7, 13.5] & 18.88 & $< .001$ \\
Enacted / Self-Directed  & 290.18 & [190.05, 443.07] & +26.1 [25.0, 27.3] & 31.40 & $< .001$ \\
Directed / Self-Directed & 134.36 & [87.84, 205.53]  & +14.0 [13.0, 15.0] & 27.02 & $< .001$ \\
\bottomrule
\end{tabular*}

\begin{tablenotes}[flushleft]
\scriptsize
\item \textit{Note.} OR = odds ratio; pp = percentage points. Differences are calculated as the first-named condition minus the second. Confidence intervals in Panel A are pointwise; confidence intervals and $p$-values in Panel C are Tukey-adjusted.
\end{tablenotes}

\end{threeparttable}
\end{minipage}
\hfill
\begin{minipage}[t]{0.36\textwidth}
\vspace{0pt}
\centering

\vspace{1.8em}

\includegraphics[width=0.95\linewidth]{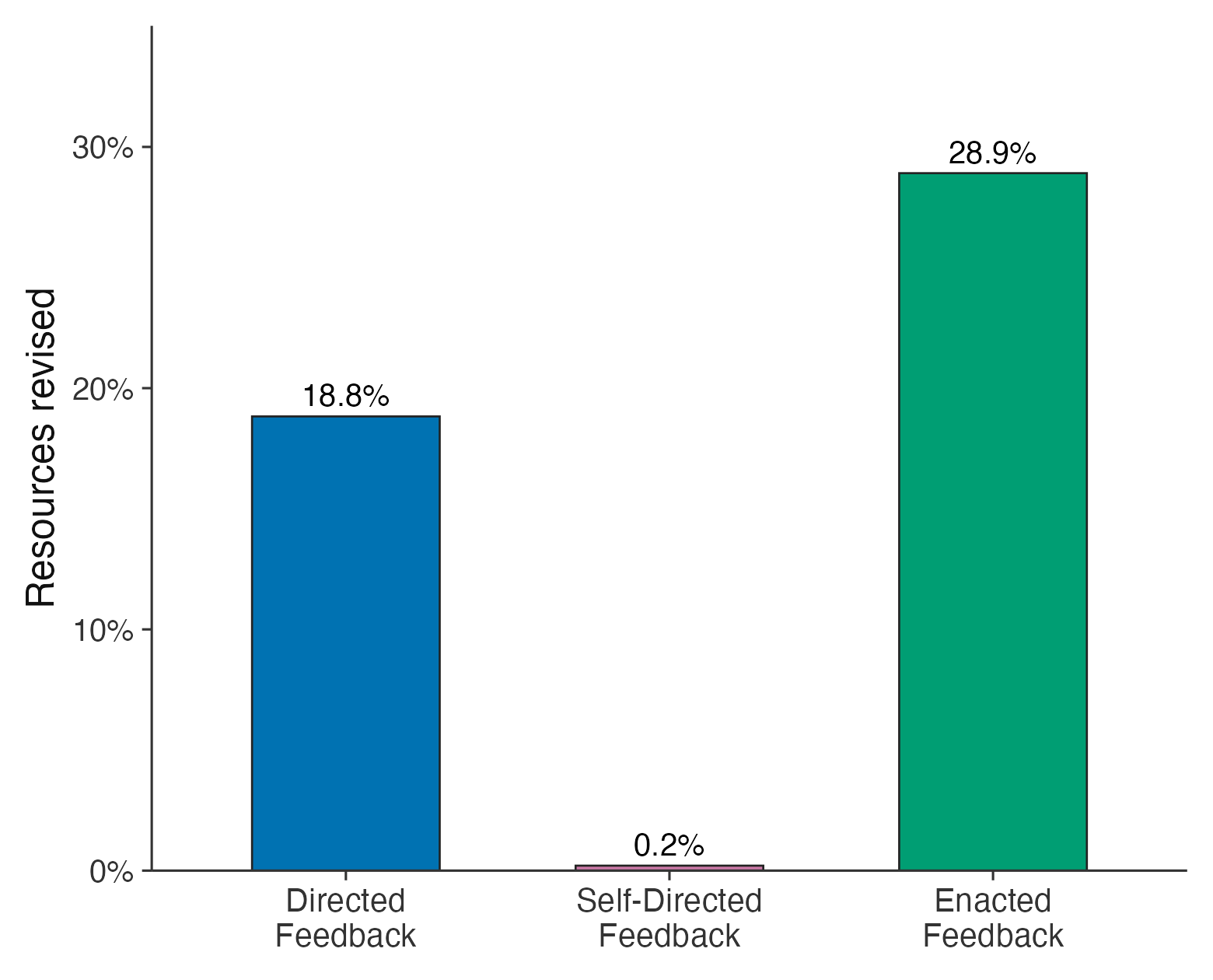}

\vspace{0.35em}

\refstepcounter{figure}
\label{fig:rq1_uptake}
\parbox{0.95\linewidth}{
\centering
\small\textbf{Figure~\thefigure:} RQ1: Workflow-specific uptake
}

\end{minipage}

\end{figure}

Table~\ref{tab:rq1_uptake} and Figure~\ref{fig:rq1_uptake} summarise uptake across the three conditions. A likelihood-ratio test showed a statistically significant overall effect of condition on  uptake, $\chi^2(2) = 5115.97$, $p < .001$. Observed uptake was highest in \textit{Enacted Feedback}, where 6,165 of 21,323 resources showed a downstream edit (28.9\%), followed by \textit{Directed Feedback}, where 2,716 of 14,425 resources showed an immediate edit (18.8\%), and \textit{Self-Directed Feedback}, where 32 of 15,548 resources showed an edit after requesting AI assistance (0.2\%). Model-estimated probabilities followed the same pattern: \textit{Enacted Feedback}, 26.2\% (95\% CI [25.3\%, 27.2\%]); \textit{Directed Feedback}, 14.1\% (95\% CI [13.3\%, 15.0\%]); and \textit{Self-Directed Feedback}, 0.1\% (95\% CI [0.1\%, 0.2\%]).

Tukey-adjusted pairwise comparisons indicated that the odds of uptake were higher in
\textit{Enacted Feedback} than in \textit{Directed Feedback}, $OR = 2.16$,
simultaneous 95\% CI [1.96, 2.38], corresponding to a model-estimated probability difference of 12.1 percentage points, simultaneous 95\% CI [10.7, 13.5], $p < .001$.
\textit{Directed Feedback} also showed higher odds of uptake than
\textit{Self-Directed Feedback}, $OR = 134.36$,
simultaneous 95\% CI [87.84, 205.53], corresponding to a model-estimated probability difference of 14.0 percentage points, simultaneous 95\% CI [13.0, 15.0], $p < .001$. \textit{Enacted Feedback} showed higher odds of uptake than
\textit{Self-Directed Feedback}, $OR = 290.18$,
simultaneous 95\% CI [190.05, 443.07], corresponding to a model-estimated probability difference of 26.1 percentage points, simultaneous 95\% CI [25.0, 27.3], $p < .001$.
These findings indicate that uptake was most frequent when feedback comments
were embedded in the \textit{Enacted Feedback} workflow, followed by static \textit{Directed
Feedback}, with very limited uptake when AI assistance was optional and
student-initiated: \textit{Enacted Feedback} $>$ \textit{Directed Feedback}
$>$ \textit{Self-Directed Feedback}.

\paragraph{\textbf{Revision counts.}}

\begin{figure}[!htbp]
\centering

\begin{minipage}[t]{0.61\textwidth}
\vspace{0pt}
\centering

\refstepcounter{table}
\label{tab:rq1_revisions}
{\small\textbf{Table~\thetable:} RQ1: Revision counts\par}

\vspace{0.35em}

\begin{threeparttable}
\centering
\tiny
\setlength{\tabcolsep}{1.8pt}
\renewcommand{\arraystretch}{1.08}

\textbf{Panel A. Descriptive statistics and model-estimated revision counts}

\vspace{0.35em}
\begin{tabular*}{\linewidth}{@{\extracolsep{\fill}}lrrrl@{}}
\toprule
\textbf{Condition} &
\textbf{Resources} &
\textbf{Students} &
\textbf{Est.\ revisions} &
\textbf{95\% CI} \\
\midrule
\textit{Directed Feedback}      & 14{,}425 & 3{,}723 & 0.239 & [0.222, 0.257] \\
\textit{Self-Directed Feedback} & 15{,}548 & 3{,}951 & 0.0034 & [0.0027, 0.0042] \\
\textit{Enacted Feedback}       & 21{,}323 & 5{,}363 & 0.602 & [0.572, 0.635] \\
\bottomrule
\end{tabular*}

\vspace{0.75em}
\textbf{Panel B. Omnibus likelihood-ratio test}

\vspace{0.35em}
\begin{tabular*}{\linewidth}{@{\extracolsep{\fill}}llrrr@{}}
\toprule
\textbf{Outcome} & \textbf{Model} & \textbf{$\chi^2$} & \textbf{df} & \textbf{$p$} \\
\midrule
Revision count & Negative-binomial GLMM & 5766.41 & 2 & $< .001$ \\
\bottomrule
\end{tabular*}

\vspace{0.75em}
\textbf{Panel C. Tukey-adjusted post hoc contrasts}

\vspace{0.35em}
\begin{tabular*}{\linewidth}{@{\extracolsep{\fill}}llllrr@{}}
\toprule
\textbf{Contrast} & \textbf{IRR} & \textbf{IRR 95\% CI} & \textbf{Difference (95\% CI)} & \textbf{$z$} & \textbf{$p_{\mathrm{adj}}$} \\
\midrule
Enacted / Directed       & 2.52   & [2.29, 2.77]     & +0.363 [0.325, 0.402] & 22.72 & $< .001$ \\
Enacted / Self-Directed  & 177.67 & [136.99, 230.41] & +0.599 [0.561, 0.637] & 46.70 & $< .001$ \\
Directed / Self-Directed & 70.46  & [54.27, 91.49]   & +0.236 [0.215, 0.256] & 38.19 & $< .001$ \\
\bottomrule
\end{tabular*}

\begin{tablenotes}[flushleft]
\scriptsize
\item \textit{Note.} IRR = incidence-rate ratio. Differences are calculated as the first-named condition minus the second. Confidence intervals in Panel A are pointwise; confidence intervals and $p$-values in Panel C are Tukey-adjusted.
\end{tablenotes}

\end{threeparttable}
\end{minipage}
\hfill
\begin{minipage}[t]{0.36\textwidth}
\vspace{0pt}
\centering

\vspace{1.8em}

\includegraphics[width=0.95\linewidth]{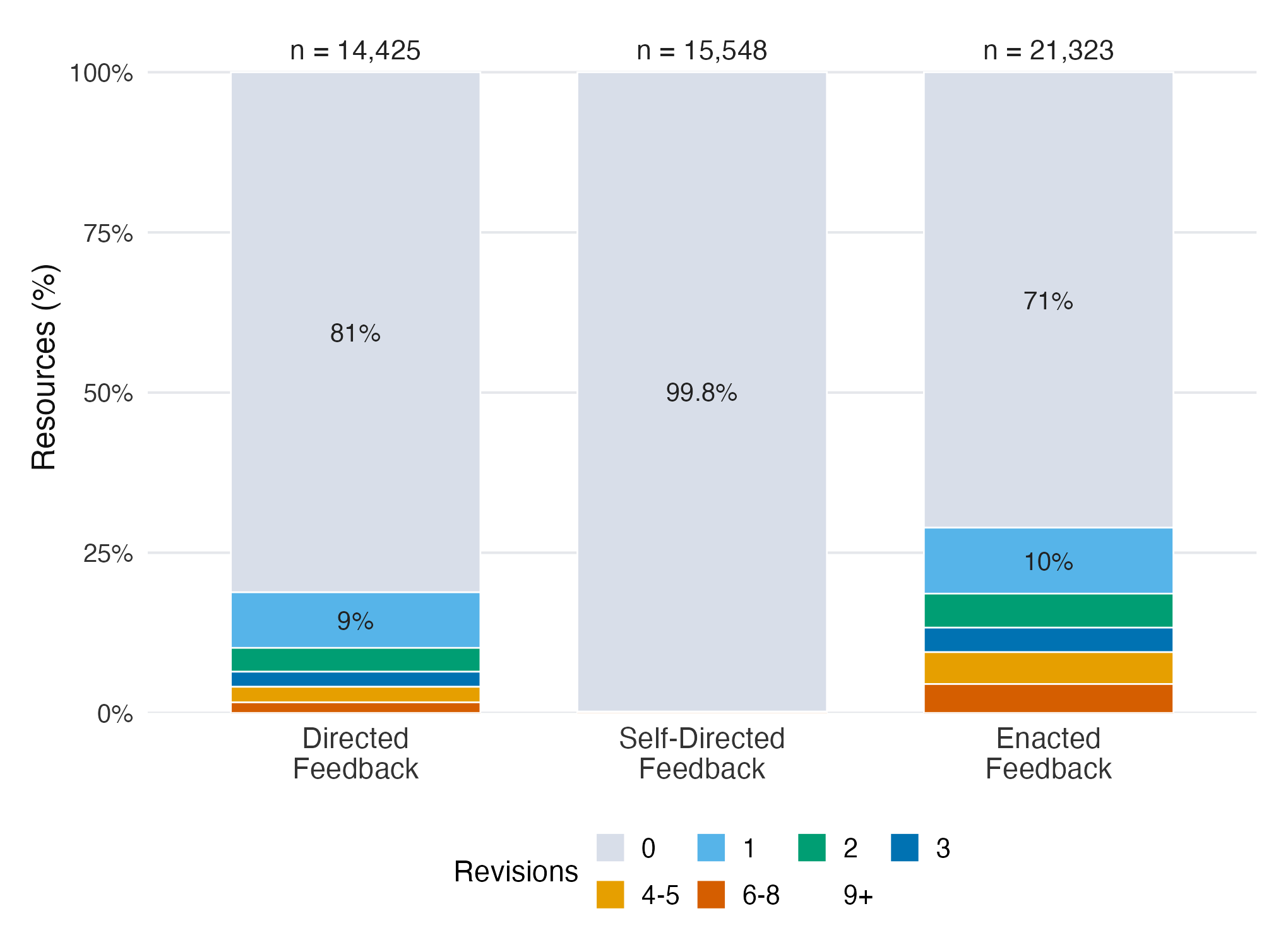}

\vspace{0.35em}

\refstepcounter{figure}
\label{fig:rq1_revision_counts}
\parbox{0.95\linewidth}{
\centering
\small\textbf{Figure~\thefigure:} RQ1: Revision counts
}

\end{minipage}

\end{figure}

Figure~\ref{fig:rq1_revision_counts} shows the number of revisions made after the relevant workflow state, and the corresponding model estimates are reported in Table~\ref{tab:rq1_revisions}. A likelihood-ratio test showed a statistically significant overall effect of condition on revision counts, $\chi^2(2) = 5766.41$, $p < .001$. Mean revision counts after capping were highest in \textit{Enacted Feedback} ($M = 0.87$, $SD = 1.82$), followed by \textit{Directed Feedback} ($M = 0.46$, $SD = 1.26$), and \textit{Self-Directed Feedback} ($M = 0.01$, $SD = 0.24$). Observed medians were 0 in all three conditions, indicating that revision behaviour remained sparse despite differences in the upper part of the distribution.

Model-estimated revision counts followed the same ordering: \textit{Enacted Feedback}, 0.602 revisions per resource (95\% CI [0.572, 0.635]); \textit{Directed Feedback}, 0.239 (95\% CI [0.222, 0.257]); and \textit{Self-Directed Feedback}, 0.0034 (95\% CI [0.0027, 0.0042]). Tukey-adjusted pairwise comparisons showed that \textit{Enacted Feedback} exceeded \textit{Directed Feedback} by 0.363 revisions per resource, simultaneous 95\% CI [0.325, 0.402], $IRR = 2.52$, simultaneous 95\% CI [2.29, 2.77], $p < .001$. \textit{Enacted Feedback} exceeded \textit{Self-Directed Feedback} by 0.599 revisions per resource, simultaneous 95\% CI [0.561, 0.637], $IRR = 177.67$, simultaneous 95\% CI [136.99, 230.41], $p < .001$. \textit{Directed Feedback} exceeded \textit{Self-Directed Feedback} by 0.236 revisions per resource, simultaneous 95\% CI [0.215, 0.256], $IRR = 70.46$, simultaneous 95\% CI [54.27, 91.49], $p < .001$. The very large IRRs for contrasts involving \textit{Self-Directed Feedback} reflect its near-zero expected count; the absolute differences provide a more direct indication of the magnitude on the count scale.

\begin{figure}[!htbp]
    \centering
    \begin{subfigure}[t]{0.32\linewidth}
        \centering
        \includegraphics[width=\linewidth]{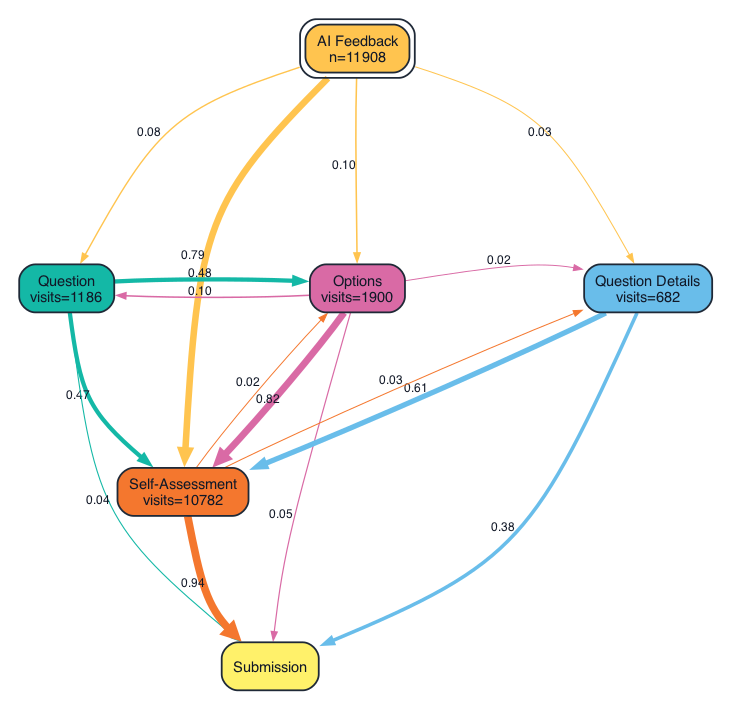}
        \caption{\textit{Directed Feedback}}
        \label{fig:fomm-directed}
    \end{subfigure}
    \hfill
    \begin{subfigure}[t]{0.32\linewidth}
        \centering
        \includegraphics[width=\linewidth]{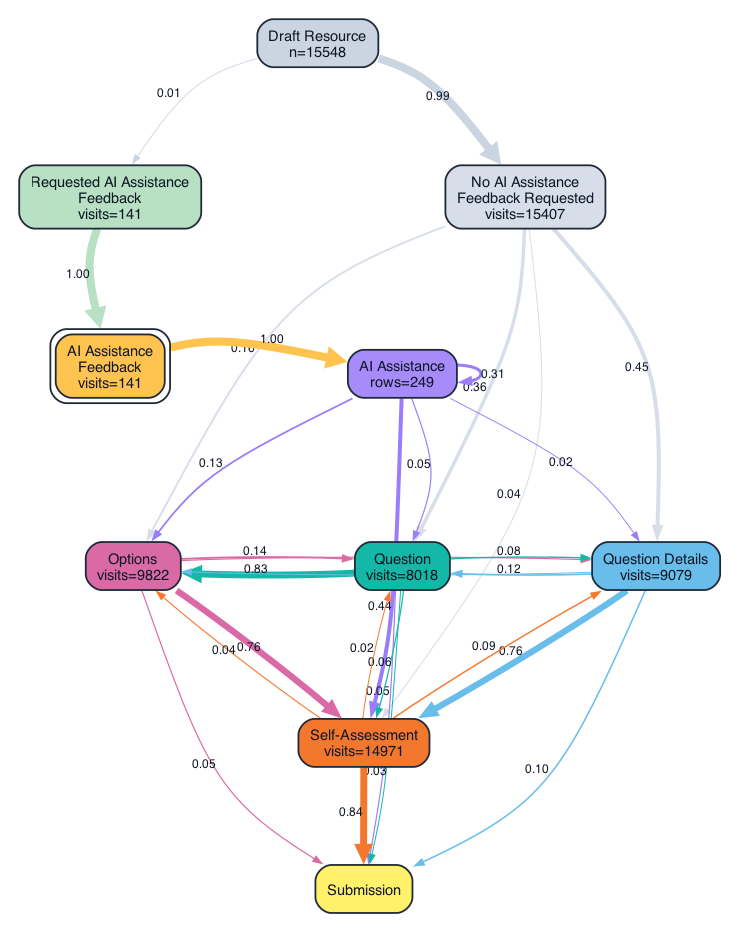}
        \caption{\textit{Self-Directed Feedback}}
        \label{fig:fomm-self-directed}
    \end{subfigure}
    \hfill
    \begin{subfigure}[t]{0.32\linewidth}
        \centering
        \includegraphics[width=\linewidth]{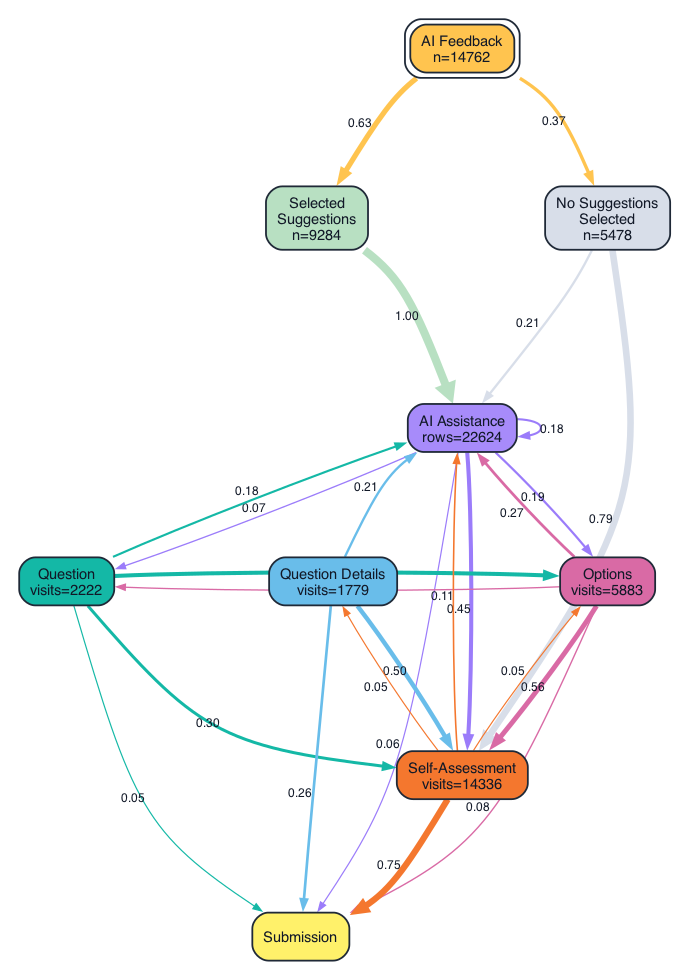}
        \caption{\textit{Enacted Feedback}}
        \label{fig:fomm-enacted}
    \end{subfigure}
    \caption{RQ1: First-order Markov model visualisations of students' behavioural transitions across the three feedback workflow conditions.}
    \label{fig:rq1_fomm}
\end{figure}

\paragraph{\textbf{Event-flow transitions.}}
Figure~\ref{fig:rq1_fomm} displays event-flow graphs from the relevant feedback or drafting state onward. Node sizes represent visit frequencies, and edge thickness represents transition probabilities. In \textit{Directed Feedback}, 78.6\% of transitions from \texttt{AI Feedback} moved directly to \texttt{Self-Assessment}, while a smaller proportion moved immediately to editing states (8.2\% to \texttt{Question} and 10.0\% to \texttt{Options}). In \textit{Self-Directed Feedback}, only 1.0\% of drafted resources moved from \texttt{Draft Resource} to \texttt{AI Assistance} after support was requested, while 99.0\% did not enter the optional AI assistance pathway; the uptake count of 32 reported above reflects the small subset of these resources that went on to edit after passing through one or more intermediate AI assistance states. In \textit{Enacted Feedback}, 63.0\% of resources moved from \texttt{AI Feedback} to selected suggestions and 37.0\% proceeded without selecting suggestions, with 28.9\% showing downstream editing after feedback comments. Overall, these findings indicate that the enacted workflow produced the strongest revision-oriented engagement, while optional AI assistance was rarely used: \textit{Enacted Feedback} $>$ \textit{Directed Feedback} $>$ \textit{Self-Directed Feedback}.

\subsection{RQ2: Self-assessment confidence}

\begin{figure}[!htbp]
\centering

\begin{minipage}[t]{0.61\textwidth}
\vspace{0pt}
\centering

\refstepcounter{table}
\label{tab:rq2_confidence}
{\small\textbf{Table~\thetable:} RQ2: Self-assessment confidence\par}

\vspace{0.35em}

\begin{threeparttable}
\centering
\tiny
\setlength{\tabcolsep}{1.8pt}
\renewcommand{\arraystretch}{1.08}

\textbf{Panel A. Descriptive statistics and estimated marginal means}

\vspace{0.35em}
\begin{tabular*}{\linewidth}{@{\extracolsep{\fill}}lrrrrrrl@{}}
\toprule
\textbf{Condition} & \textbf{Obs.} & \textbf{Students} & \textbf{$M$} & \textbf{$Mdn$} & \textbf{$SD$} & \textbf{EMM} & \textbf{EMM 95\% CI} \\
\midrule
\textit{Directed Feedback}      & 14{,}004 & 3{,}709 & 4.13 & 4.00 & 0.72 & 4.13 & [4.11, 4.15] \\
\textit{Self-Directed Feedback} & 4{,}768  & 2{,}841 & 4.02 & 4.00 & 0.76 & 4.03 & [4.00, 4.05] \\
\textit{Enacted Feedback}       & 17{,}568 & 5{,}312 & 4.19 & 4.00 & 0.71 & 4.20 & [4.18, 4.21] \\
\bottomrule
\end{tabular*}

\vspace{0.75em}
\textbf{Panel B. Omnibus likelihood-ratio test}

\vspace{0.35em}
\begin{tabular*}{\linewidth}{@{\extracolsep{\fill}}llrrr@{}}
\toprule
\textbf{Outcome} & \textbf{Model} & \textbf{$\chi^2$} & \textbf{df} & \textbf{$p$} \\
\midrule
Confidence & CLMM (logit link) & 139.06 & 2 & $< .001$ \\
\bottomrule
\end{tabular*}

\vspace{0.75em}
\textbf{Panel C. Tukey-adjusted post hoc contrasts}

\vspace{0.35em}
\begin{tabular*}{\linewidth}{@{\extracolsep{\fill}}llllrr@{}}
\toprule
\textbf{Contrast} & \textbf{Cumulative OR} & \textbf{OR 95\% CI} & \textbf{Expected-rating difference (95\% CI)} & \textbf{$z$} & \textbf{$p_{\mathrm{adj}}$} \\
\midrule
Enacted / Directed       & 1.41 & [1.21, 1.65] & +0.070 [0.039, 0.101] & 5.21  & $< .001$ \\
Enacted / Self-Directed  & 2.44 & [2.04, 2.91] & +0.170 [0.136, 0.204] & 11.75 & $< .001$ \\
Directed / Self-Directed & 1.73 & [1.43, 2.08] & +0.100 [0.066, 0.134] & 6.84  & $< .001$ \\
\bottomrule
\end{tabular*}

\begin{tablenotes}[flushleft]
\scriptsize
\item \textit{Note.} EMM = estimated marginal mean; OR = cumulative odds ratio. EMMs and differences are reported on the 1--5 confidence scale. Differences are calculated as the first-named condition minus the second. Confidence intervals in Panel A are pointwise; confidence intervals and $p$-values in Panel C are Tukey-adjusted.
\end{tablenotes}

\end{threeparttable}
\end{minipage}
\hfill
\begin{minipage}[t]{0.36\textwidth}
\vspace{0pt}
\centering

\vspace{1.8em}

\includegraphics[width=0.95\linewidth]{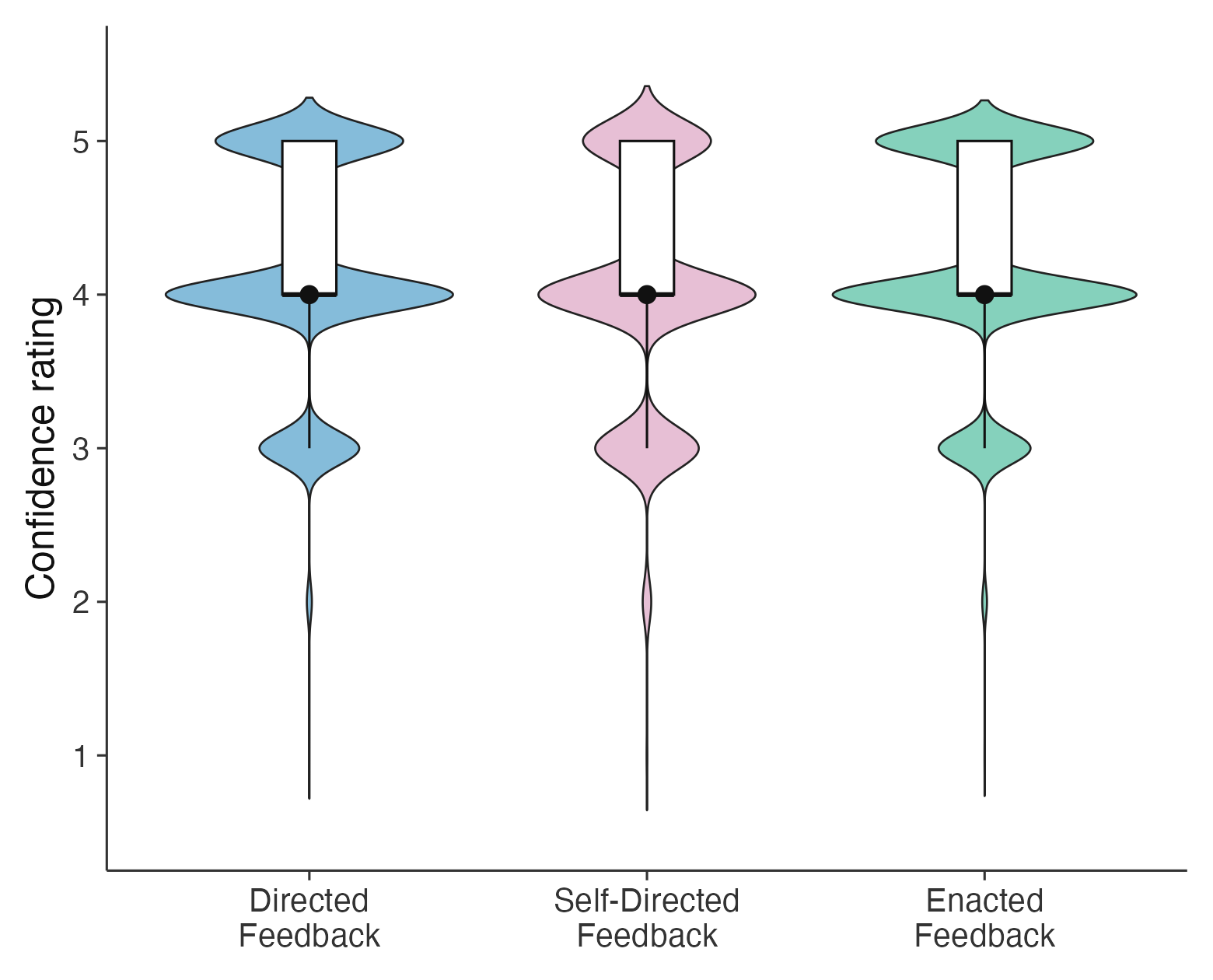}

\vspace{0.35em}

\refstepcounter{figure}
\label{fig:rq2_confidence}
\parbox{0.95\linewidth}{
\centering
\small\textbf{Figure~\thefigure:} RQ2: Self-assessment confidence
}

\end{minipage}

\end{figure}

Table~\ref{tab:rq2_confidence} and Figure~\ref{fig:rq2_confidence} summarise self-assessment confidence across conditions. A likelihood-ratio test showed a statistically significant overall effect of condition on self-assessment confidence, $\chi^2(2) = 139.06$, $p < .001$. Observed confidence was high in all conditions: \textit{Enacted Feedback} ($M = 4.19$, $SD = 0.71$), \textit{Directed Feedback} ($M = 4.13$, $SD = 0.72$), and \textit{Self-Directed Feedback} ($M = 4.02$, $SD = 0.76$). Observed medians were identical across conditions ($Mdn = 4$).

EMMs on the expected rating scale indicated the same descriptive ordering: \textit{Enacted Feedback}, 4.20 (95\% CI [4.18, 4.21]); \textit{Directed Feedback}, 4.13 (95\% CI [4.11, 4.15]); and \textit{Self-Directed Feedback}, 4.03 (95\% CI [4.00, 4.05]). Tukey-adjusted pairwise comparisons showed that \textit{Enacted Feedback} was associated with higher cumulative odds of reporting higher confidence than \textit{Directed Feedback}, $OR = 1.41$, simultaneous 95\% CI [1.21, 1.65], with an expected-rating difference of 0.070, simultaneous 95\% CI [0.039, 0.101], $p < .001$. \textit{Enacted Feedback} was also associated with higher cumulative odds of reporting higher confidence than \textit{Self-Directed Feedback}, $OR = 2.44$, simultaneous 95\% CI [2.04, 2.91], with an expected-rating difference of 0.170, simultaneous 95\% CI [0.136, 0.204], $p < .001$. Confidence was also higher in \textit{Directed Feedback} than in \textit{Self-Directed Feedback}, $OR = 1.73$, simultaneous 95\% CI [1.43, 2.08], with an expected-rating difference of 0.100, simultaneous 95\% CI [0.066, 0.134], $p < .001$. These findings indicate that confidence ratings followed a statistically reliable ordering, but the differences on the expected 1--5 rating scale were modest: \textit{Enacted Feedback} $>$ \textit{Directed Feedback} $>$ \textit{Self-Directed Feedback}.

\subsection{RQ3: Submitted-work quality}

\begin{figure}[!htbp]
\centering

\begin{minipage}[t]{0.61\textwidth}
\vspace{0pt}
\centering

\refstepcounter{table}
\label{tab:rq3_moderation}
{\small\textbf{Table~\thetable:} RQ3: Submitted-work quality\par}

\vspace{0.35em}

\begin{threeparttable}
\centering
\tiny
\setlength{\tabcolsep}{1.8pt}
\renewcommand{\arraystretch}{1.08}

\textbf{Panel A. Descriptive statistics and estimated marginal means}

\vspace{0.35em}
\begin{tabular*}{\linewidth}{@{\extracolsep{\fill}}lrrrrrrl@{}}
\toprule
\textbf{Condition} & \textbf{Obs.} & \textbf{Students} & \textbf{$M$} & \textbf{$Mdn$} & \textbf{$SD$} & \textbf{EMM} & \textbf{EMM 95\% CI} \\
\midrule
\textit{Directed Feedback}      & 13{,}821 & 3{,}713 & 4.13 & 4.20 & 0.55 & 4.191 & [4.177, 4.205] \\
\textit{Self-Directed Feedback} & 14{,}923 & 3{,}924 & 4.18 & 4.20 & 0.47 & 4.244 & [4.231, 4.257] \\
\textit{Enacted Feedback}       & 16{,}417 & 5{,}290 & 4.22 & 4.30 & 0.51 & 4.328 & [4.317, 4.338] \\
\bottomrule
\end{tabular*}

\vspace{0.75em}
\textbf{Panel B. Omnibus likelihood-ratio test}

\vspace{0.35em}
\begin{tabular*}{\linewidth}{@{\extracolsep{\fill}}llrrr@{}}
\toprule
\textbf{Outcome} & \textbf{Model} & \textbf{$\chi^2$} & \textbf{df} & \textbf{$p$} \\
\midrule
Moderation outcome & Beta GLMM (logit link) & 251.30 & 2 & $< .001$ \\
\bottomrule
\end{tabular*}

\vspace{0.75em}
\textbf{Panel C. Tukey-adjusted post hoc contrasts}

\vspace{0.35em}
\begin{tabular*}{\linewidth}{@{\extracolsep{\fill}}llllrr@{}}
\toprule
\textbf{Contrast} & \textbf{EMM difference} & \textbf{95\% CI} & \textbf{Expected-proportion OR (95\% CI)} & \textbf{$z$} & \textbf{$p_{\mathrm{adj}}$} \\
\midrule
Enacted / Directed       & +0.137 & [0.116, 0.158] & 1.24 [1.20, 1.28] & 15.58 & $< .001$ \\
Enacted / Self-Directed  & +0.083 & [0.063, 0.103] & 1.15 [1.11, 1.18] & 9.89  & $< .001$ \\
Self-Directed / Directed & +0.054 & [0.031, 0.076] & 1.08 [1.05, 1.12] & 5.59  & $< .001$ \\
\bottomrule
\end{tabular*}

\begin{tablenotes}[flushleft]
\scriptsize
\item \textit{Note.} EMM = estimated marginal mean; OR = expected-proportion odds ratio. EMMs and differences are reported on the original 0--5 submitted-work quality scale. Differences are calculated as the first-named condition minus the second. Confidence intervals in Panel A are pointwise; confidence intervals and $p$-values in Panel C are Tukey-adjusted.
\end{tablenotes}

\end{threeparttable}
\end{minipage}
\hfill
\begin{minipage}[t]{0.36\textwidth}
\vspace{0pt}
\centering

\vspace{1.8em}

\includegraphics[width=0.95\linewidth]{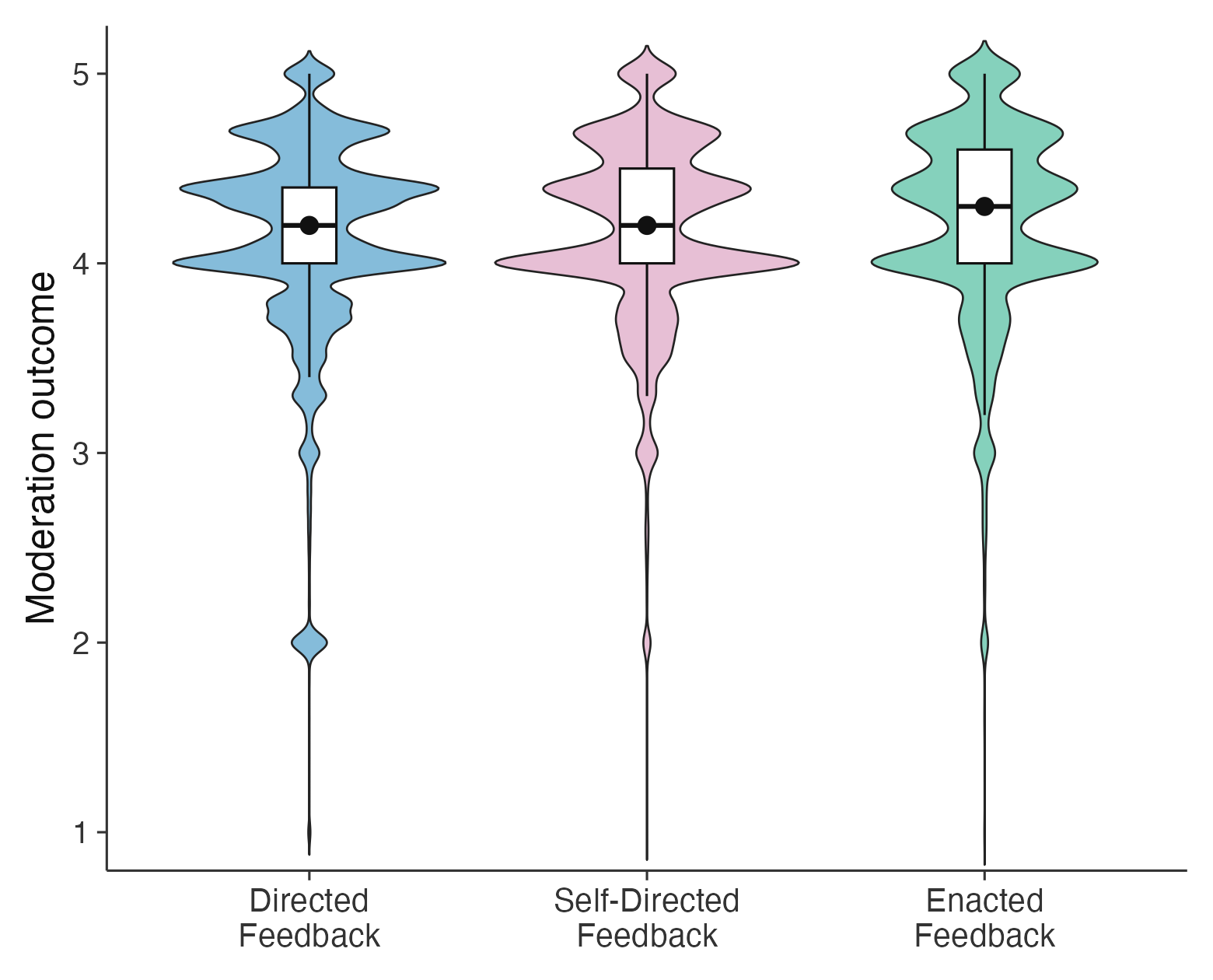}

\vspace{0.35em}

\refstepcounter{figure}
\label{fig:rq3_moderation}
\parbox{0.95\linewidth}{
\centering
\small\textbf{Figure~\thefigure:} RQ3: Submitted-work quality
}

\end{minipage}

\end{figure}

Table~\ref{tab:rq3_moderation} and Figure~\ref{fig:rq3_moderation} summarise submitted-work quality across conditions, operationalised using peer moderation outcome scores. A likelihood-ratio test showed a statistically significant overall effect of condition on submitted-work quality, $\chi^2(2) = 251.30$, $p < .001$. Observed moderation outcomes were high across conditions, with \textit{Enacted Feedback} showing the highest mean score ($M = 4.22$, $SD = 0.51$), followed by \textit{Self-Directed Feedback} ($M = 4.18$, $SD = 0.47$), and \textit{Directed Feedback} ($M = 4.13$, $SD = 0.55$). Observed medians were 4.20 for \textit{Directed Feedback}, 4.20 for \textit{Self-Directed Feedback}, and 4.30 for \textit{Enacted Feedback}.

Back-transformed EMMs on the original 0--5 scale indicated that \textit{Enacted Feedback} had the highest estimated moderation score, 4.328 (95\% CI [4.317, 4.338]), followed by \textit{Self-Directed Feedback}, 4.244 (95\% CI [4.231, 4.257]), and \textit{Directed Feedback}, 4.191 (95\% CI [4.177, 4.205]). Tukey-adjusted pairwise comparisons showed that \textit{Enacted Feedback} exceeded \textit{Directed Feedback} by 0.137 points, simultaneous 95\% CI [0.116, 0.158], with an expected-proportion odds ratio of 1.24, simultaneous 95\% CI [1.20, 1.28], $p < .001$. \textit{Enacted Feedback} exceeded \textit{Self-Directed Feedback} by 0.083 points, simultaneous 95\% CI [0.063, 0.103], with an expected-proportion odds ratio of 1.15, simultaneous 95\% CI [1.11, 1.18], $p < .001$. \textit{Self-Directed Feedback} exceeded \textit{Directed Feedback} by 0.054 points, simultaneous 95\% CI [0.031, 0.076], with an expected-proportion odds ratio of 1.08, simultaneous 95\% CI [1.05, 1.12], $p < .001$. These response-scale differences indicate a statistically reliable ordering with small differences in submitted-work quality: \textit{Enacted Feedback} $>$ \textit{Self-Directed Feedback} $>$ \textit{Directed Feedback}.

\section{Discussion}

This study examined how different AI-mediated feedback workflows were associated with students' behavioural engagement, self-assessment confidence, and submitted-work quality. Rather than treating AI-generated feedback comments as static products, the study compared three ways of organising students' encounters with AI-generated feedback comments or optional AI assistance: \textit{Directed Feedback}, \textit{Self-Directed Feedback}, and \textit{Enacted Feedback}. Overall, the findings suggest that the educational value of AI-generated feedback comments depends not only on their availability or quality, but also on how students are guided to interpret, prioritise, discuss, and act on them. This supports contemporary feedback scholarship, which views feedback as an active process of learner sense-making and action rather than the passive transmission of information \citep{carless2023teacher,nicol2021power,winstone2017supporting,tai2018developing,chong2021reconsidering}.

\subsection{RQ1: Uptake, revision counts, and event-flow transitions}

The first research question examined how the three workflows differed in students' behavioural engagement. The findings, summarised in Tables~\ref{tab:rq1_uptake}--\ref{tab:rq1_revisions} and Figures~\ref{fig:rq1_uptake}--\ref{fig:rq1_revision_counts}, showed that \textit{Enacted Feedback} was associated with consistently stronger behavioural engagement than both \textit{Directed Feedback} and \textit{Self-Directed Feedback}. Students in the \textit{Enacted Feedback} condition showed higher uptake, made more subsequent revisions, and moved through more revision-oriented event flows. \textit{Directed Feedback} was associated with some uptake and revision activity, but at a lower level than \textit{Enacted Feedback}, whereas students in \textit{Self-Directed Feedback} rarely initiated the optional AI pathway.

The limited use of the \textit{Self-Directed Feedback} pathway suggests that making optional AI assistance available was insufficient to initiate feedback engagement for most students. Accessing this support required students to recognise a need for assistance, decide to seek it, and formulate an appropriate request. These activities are more accurately understood as SRL processes involving monitoring, help-seeking, and strategy selection than as evaluative judgement, which becomes relevant once feedback information is available for appraisal \citep{zimmerman2002becoming, panadero2017review, panadero2019using}. Because most students did not enter the pathway, the primary difficulty appears to have arisen before an AI-mediated feedback encounter occurred. This interpretation is consistent with evidence that engagement with GenAI-supported feedback functions can remain modest when their outputs do not align with students' expectations or are perceived as unnecessary \citep{jin2025students}. However, students' reasons for not initiating assistance were not measured directly, so the explanation remains theoretically grounded rather than empirically confirmed.

The stronger engagement associated with \textit{Enacted Feedback} indicates that workflow structure can scaffold the regulatory processes involved in feedback enactment. The workflow made selection, evaluation, AI-supported dialogue, and revision explicit parts of the encounter. Evaluative judgement supported students in assessing which suggestions were relevant, while movement from those judgements towards revision involved broader SRL processes of decision-making and action \citep{panadero2019using}. By prompting these activities, the workflow reduced students' dependence on independently initiating and coordinating them. This interpretation aligns with formative-feedback research that emphasises active evaluation and action rather than the passive receipt of comments \citep{nicol2006formative, malecka2022eliciting}.

\textit{Directed Feedback} occupied an intermediate position. Presenting AI-generated feedback comments directly removed the need for students to initiate the optional AI-assistance pathway. However, students still needed to determine independently which comments were relevant and how they should be used. The difference between \textit{Directed Feedback} and \textit{Enacted Feedback} therefore suggests that making comments available can initiate some engagement, but more consistent enactment may require support for the regulatory processes that follow the presentation of comments \citep{carless2018development, wood2021dialogic, wood2023enabling}.

\subsection{RQ2: Self-assessment confidence}

The second research question examined whether the workflows differed in students' self-assessment confidence. The findings, summarised in Table~\ref{tab:rq2_confidence} and Figure~\ref{fig:rq2_confidence}, showed a statistically reliable ordering: confidence was highest in \textit{Enacted Feedback}, followed by \textit{Directed Feedback}, and then \textit{Self-Directed Feedback}. \textit{Enacted Feedback} was significantly higher than both other conditions, while \textit{Directed Feedback} was significantly higher than \textit{Self-Directed Feedback}. This ordering should be interpreted in light of students' exposure to feedback information: students in the first two conditions encountered AI-generated feedback comments, whereas most students in \textit{Self-Directed Feedback} did not initiate the optional pathway and therefore did not see or use such comments.

The difference between \textit{Enacted Feedback} and \textit{Directed Feedback} provides the clearest indication of how workflow structure may relate to confidence because students in both conditions encountered AI-generated feedback comments. The higher confidence associated with \textit{Enacted Feedback} may reflect the clearer basis for self-assessment created by its structured encounter. By directing attention towards specific aspects of students' work and supporting clarification, the workflow may have reduced uncertainty about the quality of their submissions and supported metacognitive monitoring \citep{zimmerman2002becoming, panadero2017review}. This interpretation is consistent with evidence that feedback-literacy interventions can strengthen students’ confidence and agency in feedback processes \citep{Little02012024} . Although these constructs differ from confidence in a specific self-assessment, they provide relevant evidence that structured feedback encounters can support students' perceived competence.

The intermediate confidence observed in \textit{Directed Feedback} may reflect the availability of external information against which students could consider the quality of their work. Feedback comments can provide reference points that enable students to compare their current performance with task goals or standards, thereby informing self-monitoring \citep{nicol2006formative, nicol2021power}. The absence of a structured process for evaluating and clarifying those comments may explain why confidence remained lower than in \textit{Enacted Feedback}.

The lower confidence observed in \textit{Self-Directed Feedback} should not be interpreted as a negative response to AI-generated feedback comments because most students did not encounter such comments. Without external feedback information, students had fewer opportunities to test and refine their judgements through comparison, potentially leaving greater uncertainty in their self-assessments. This interpretation aligns with comparison-based accounts in which students generate internal feedback by comparing their work with external information, criteria, or other reference points \citep{nicol2021power}. The absence of this comparison opportunity may therefore have contributed to the lower confidence observed in this condition, although pre-existing differences in confidence or regulatory capability across cohorts cannot be excluded.

Confidence should also be distinguished from self-assessment accuracy. A higher confidence rating does not necessarily indicate that students judged the quality of their work more accurately, just as lower confidence may reflect either uncertainty or more critical monitoring. Establishing confidence calibration would require examining whether students' confidence corresponded with the accuracy of their self-assessments relative to independently assessed submitted-work quality \citep{boud2015calibration}.

\subsection{RQ3: Submitted-work quality}

The third research question examined whether the three AI-mediated feedback workflows differed in submitted-work quality, measured using peer moderation outcome scores. As shown in Table~\ref{tab:rq3_moderation} and Figure~\ref{fig:rq3_moderation}, \textit{Enacted Feedback} was associated with significantly higher submitted-work quality than both \textit{Directed Feedback} and \textit{Self-Directed Feedback}. \textit{Self-Directed Feedback} was also associated with significantly higher submitted-work quality than \textit{Directed Feedback}. However, this difference was small in practical terms and cannot be attributed to optional AI assistance because most students in \textit{Self-Directed Feedback} did not  encounter AI-generated feedback comments.

The comparison between \textit{Enacted Feedback} and \textit{Directed Feedback} provides the clearest evidence concerning workflow structure because students in both conditions encountered AI-generated feedback comments. The higher submitted-work quality associated with \textit{Enacted Feedback} suggests that comments were more educationally consequential when the surrounding process scaffolded their evaluation and enactment. Evaluative judgement supported the assessment of suggestion relevance, while translating those judgements into changes involved the broader SRL processes of monitoring, strategy selection, and action \citep{zimmerman2002becoming, panadero2017review, panadero2019using}. This interpretation aligns with research emphasising that feedback comments influence outcomes through the processes by which students make sense of and act on them \citep{nicol2006formative, malecka2022eliciting, wood2021dialogic}. This pattern locates the difference in the process surrounding the comments: when comments were presented without structured support for their interpretation and enactment, they were less consistently translated into improvements in submitted-work quality.

The small advantage of \textit{Self-Directed Feedback} over \textit{Directed Feedback} requires a different interpretation. Because most students did not enter the optional AI pathway, the difference does not provide evidence that optional AI assistance improved submitted-work quality. One possibility is that students in this cohort relied more successfully on their own monitoring, judgement, and self-regulatory processes when preparing their submissions \citep{zimmerman2002becoming, panadero2017review}. However, SRL was not measured directly, and the sequential-cohort design cannot exclude pre-existing differences in academic capability, motivation, confidence, or other cohort characteristics. Given the small practical effect, this difference provides limited evidence of a meaningful educational advantage and does not indicate that \textit{Directed Feedback} reduced submitted-work quality. 

The clearest finding is the advantage of \textit{Enacted Feedback}. By requiring students to select, evaluate, and act on feedback suggestions, the workflow scaffolded the feedback-literacy and self-regulatory processes needed to translate AI-generated feedback comments into revision. This aligns with research showing that feedback is more effective when students are supported to judge and enact comments \citep{malecka2022eliciting, wood2021dialogic}.

\subsection{Theoretical implications}

This study contributes to feedback literacy research by showing how the learning environment can scaffold behaviours associated with productive feedback use. Rather than assuming that students already know how to interpret and use feedback comments, the \textit{Enacted Feedback} workflow embedded structured opportunities to select suggestions, judge their relevance, enter AI-supported dialogue, and revise. This supports ecological and curriculum-oriented perspectives on feedback literacy, which argue that feedback use is shaped by task design, learning context, and opportunities for action \citep{chong2021reconsidering, winstone2022discipline}.

The study also contributes to research on AI-generated feedback comments by shifting attention from comment generation to feedback enactment. Much of this research examines whether AI can produce useful or accurate comments. The present findings suggest that this question is necessary but incomplete. Even useful AI-generated feedback comments may have limited impact if students are not supported to understand, evaluate, and act on them. This consideration is particularly relevant because students' trust in AI, perceptions of feedback source, and AI literacy can influence whether they accept or act on AI-generated feedback comments \citep{guardia2026human, nazaretsky2026gives}.

\subsection{Practical implications}

For educators, the findings suggest that AI-generated feedback comments should not be implemented in isolation. Students need structured opportunities to decide which suggestions are relevant, clarify their meaning, and translate them into revision. This implies that AI-mediated feedback workflows should include prompts for selection, prioritisation, explanation, and revision planning.

For learning designers and platform developers, the findings highlight the relevance of workflow structure. Optional AI assistance may be underused when it is separated from the feedback process. By contrast, embedding AI-supported dialogue directly within the feedback pathway can make support more visible, timely, and actionable. AI-mediated feedback systems should therefore be designed not merely to generate comments but to guide students through the process of using them.

\subsection{Limitations and future research}
\label{sec:limitations}

The findings reported above should be interpreted alongside several methodological constraints that bear on the strength and generalisability of the conclusions.

First, the language model used for AI-mediated support differed across the three implementation periods. The \textit{Directed Feedback} and \textit{Self-Directed Feedback} conditions used GPT-4o mini, deployed in Semester 1 and Semester 2 of 2025, while the \textit{Enacted Feedback} condition used GPT-5 mini, deployed in Semester 1 of 2026. In the two conditions that presented AI-generated feedback comments, the prompt structure, conditioning metadata, and output format were held constant, but the model version was not. This heterogeneity reflects the practical reality that the three cohorts were deployed sequentially during a period of rapid model development and is acknowledged as a confound: differences in engagement or outcomes between the \textit{Enacted Feedback} condition and the comparison conditions could reflect the workflow structure (the focal manipulation), the model version, or both. Two observations provide limited contextual reassurance but do not resolve this confound. First, the \textit{Self-Directed Feedback} condition produced near-zero uptake despite using the same model as the \textit{Directed Feedback} condition, indicating that access to a conversational model did not by itself generate engagement. Second, the moderation outcome differences were small, although this pattern does not distinguish between workflow- and model-version explanations. Future research should address the model-version confound directly by holding the model constant across conditions.

Second, the study used a quasi-experimental sequential cohort design rather than random assignment. Although the platform, task structure, moderation rubric, and assessment logic were retained across cohorts, differences between courses, instructional cohorts, and broader institutional or seasonal conditions during each semester may have contributed to the observed effects. The sequential design also means that trends in students' familiarity with AI tools across 2025--2026 cannot be ruled out as a contributor to the patterns observed in the \textit{Enacted Feedback} condition. Future research should compare the feedback conditions within the same course and semester, using random assignment where possible, thereby reducing confounding due to cohort differences.

Third, uptake was operationalised differently across the three workflows because they embedded different routing structures. In \textit{Directed Feedback}, uptake required an immediate transition from the static-feedback state to an editing state; in \textit{Self-Directed Feedback}, it required requesting AI assistance and subsequently editing; in \textit{Enacted Feedback}, it included any downstream edit after entering the scaffolded pathway. These operationalisations are faithful to the structure of each workflow, but they preclude treating uptake as an identical behaviour across conditions. The findings should therefore be interpreted as workflow-specific indicators of behavioural engagement rather than as a strict test of equivalent behaviour. Moreover, the log data establish that editing occurred but do not show that an edit incorporated a particular feedback suggestion.

Fourth, the analyses relied on platform log data. Log data provide a robust record of observable behaviour at scale, but they cannot capture students' motivations, interpretations, or decision-making processes during feedback use. Log data do not reveal why students in the \textit{Self-Directed Feedback} condition rarely engaged with the optional AI assistance, why some students in the \textit{Enacted Feedback} condition bypassed suggestion selection, or what revision strategies students applied when they did revise. Future research should triangulate trace data with interviews, think-aloud protocols, or qualitative analysis of the revisions themselves to characterise the cognitive and affective processes that mediate workflow effects.

Fifth, the study measured submitted-work quality through peer moderation outcomes at a single point in each cohort's cycle but did not directly assess longer-term learning. Whether the engagement benefits observed under \textit{Enacted Feedback} translate into durable gains in students' feedback literacy, evaluative judgement, or independent revision skill beyond the studied task remains an open empirical question. Future research should follow students across multiple authoring cycles, ideally including transfer tasks in which AI scaffolding is absent, to determine whether scaffolded engagement during early tasks supports more independent feedback use.

Sixth, the study manipulated the workflow surrounding AI-generated feedback comments while holding the prompt structure and feedback format constant. It did not vary the content, quality, or specificity of the AI-generated feedback comments themselves. The findings therefore speak to the effect of workflow structure given a reasonable baseline quality of AI-generated feedback comments, not to the interaction between comment quality and workflow design. Future research should systematically vary comment quality and workflow conditions to examine how these factors jointly shape engagement and learning outcomes.

Finally, the present study was conducted within a single platform (RiPPLE) and within a particular pedagogical context in which students author resources for peer use. The findings may not transfer cleanly to settings in which the assessment task differs in structure, stakes, or social purpose, including individual writing assessment, problem-solving in mathematics or programming, and reflective tasks. Future research should test the \textit{Enacted Feedback} workflow structure across distinct pedagogical and disciplinary contexts to establish the boundary conditions of its effects.

\section{Conclusion}

The educational value of AI-generated feedback comments appears to depend on how students are supported to use them, not simply on whether comments are provided. The \textit{Enacted Feedback} workflow was associated with higher workflow-specific uptake, more revisions, higher self-assessment confidence, and higher submitted-work quality than either comparison workflow. The magnitude of these differences varied by outcome: uptake and revision-count contrasts showed clear behavioural separation between workflows, whereas differences in expected confidence ratings and submitted-work quality were modest on their original scales. Generative AI can therefore help address the challenge of providing timely and individualised feedback comments at scale, but this capability does not by itself address the separate challenge of student engagement with feedback.

The central distinction drawn in this study is between access and enactment. Making AI-generated feedback comments available, or giving students optional access to AI assistance, was not accompanied by the level of feedback use observed when enactment was scaffolded. The optional pathway showed very limited uptake, indicating that availability alone did not lead most students to engage with AI support. The strongest engagement and outcomes were observed in the workflow that prompted students to select feedback suggestions, consider their relevance, engage in targeted AI-supported dialogue, and revise their work. Workflow design may therefore function as a pedagogical mechanism for scaffolding behaviours associated with productive feedback use, although the sequential cohort design and the difference in model version across conditions mean this interpretation requires confirmation under randomised, model-controlled conditions.

These findings have implications for the design of AI-mediated feedback systems in higher education. The educational value of such systems lies in structuring the processes through which students interpret, evaluate, and act on comments, as well as in generating those comments quickly and consistently. A system that presents high-quality feedback comments without supporting their use addresses only part of the feedback problem. The design of feedback-use processes therefore appears consequential for whether AI-generated feedback comments translate into meaningful learning activity.

Overall, this study invites a shift in how AI-generated feedback comments are understood and evaluated. Rather than focusing only on the quality of the comments an AI system produces, researchers and designers should also examine the learning processes structured around those comments. Designing for enactment positions students as active participants in judgement, dialogue, and improvement rather than as passive recipients of automated commentary. Realising the educational potential of AI-generated feedback comments will depend on carefully designed workflows that help students use those comments productively, and not on model capability alone.

\bibliographystyle{unsrtnat}
\bibliography{references}

@article{nicol2021power,
  title={The power of internal feedback: Exploiting natural comparison processes},
  author={Nicol, David},
  journal={Assessment \& Evaluation in higher education},
  volume={46},
  number={5},
  pages={756--778},
  year={2021},
  publisher={Taylor \& Francis}
}

@article{Liang2023The,title={The relationship between student interaction with generative artificial intelligence and learning achievement: serial mediating roles of self-efficacy and cognitive engagement},author={Jing Liang and Lili Wang and Jia Luo and Yufei Yan and Chao Fan},journal={Frontiers in Psychology},year={2023},volume={14},doi={10.3389/fpsyg.2023.1285392}}

@article{zhai2023systematic,
  title={A systematic review on artificial intelligence dialogue systems for enhancing English as foreign language students’ interactional competence in the university},
  author={Zhai, Chunpeng and Wibowo, Santoso},
  journal={Computers and Education: Artificial Intelligence},
  volume={4},
  pages={100134},
  year={2023},
  publisher={Elsevier}
}

@article{Quinton04072025,
author = {Jessica Quinton and Lorien Nesbitt and Johanna Bock},
title = {Enhancing the structure of feedback forms increases trustworthiness and usefulness of peer feedback},
journal = {Assessment \& Evaluation in Higher Education},
volume = {50},
number = {5},
pages = {760--774},
year = {2025},
publisher = {Routledge},
doi = {10.1080/02602938.2025.2468848}
}

@article{khosravi2026building,
  title={Building AI Companions that Prioritise Learning over Performance},
  author={Khosravi, Hassan and Gasevic, Dragan and Sadiq, Shazia and Yan, Lixiang and Lodge, Jason and Tangen, Jason and Denny, Paul and DiCerbo, Kristen and Shum, Simon Buckingham and Baker, Ryan S},
  journal={arXiv preprint arXiv:2605.04816},
  year={2026}
}

@article{Dahri2024Investigating,title={Investigating AI-based academic support acceptance and its impact on students’ performance in Malaysian and Pakistani higher education institutions},author={Nisar Ahmed Dahri and N. Yahaya and W. Al-rahmi and M. S. Vighio and Fahad Alblehai and Rahim Bux Soomro and Anna Shutaleva},journal={Education and Information Technologies},year={2024},volume={29},pages={18695 - 18744},doi={10.1007/s10639-024-12599-x}}

@article{Almogren2024Exploring,title={Exploring factors influencing the acceptance of ChatGPT in higher education: A smart education perspective},author={Abeer S. Almogren and W. Al-rahmi and Nisar Ahmed Dahri},journal={Heliyon},year={2024},volume={10},doi={10.1016/j.heliyon.2024.e31887}}

@article{Li2025The,title={The mediating effects of needs satisfaction on the relationship between teacher support and student engagement with generative artificial intelligence (GenAI) chatbots from a self-determination theory (SDT) perspective},author={Yan Li and T. Chiu},journal={Education and Information Technologies},year={2025},volume={30},pages={20051 - 20070},doi={10.1007/s10639-025-13574-w}}

@article{Bai2025Impact,title={Impact of generative AI interaction and output quality on university students’ learning outcomes: a technology-mediated and motivation-driven approach},author={Yun Bai and Shaofeng Wang},journal={Scientific Reports},year={2025},volume={15},doi={10.1038/s41598-025-08697-6}}

@article{panadero2017review,
  title={A review of self-regulated learning: Six models and four directions for research},
  author={Panadero, Ernesto},
  journal={Frontiers in psychology},
  volume={8},
  pages={422},
  year={2017},
  publisher={Frontiers Media SA}
}

@article{boud2015calibration,
author = {David Boud and Romy Lawson and Darrall G. Thompson},
title = {The calibration of student judgement through self-assessment: disruptive effects of assessment patterns},
journal = {Higher Education Research \& Development},
volume = {34},
number = {1},
pages = {45--59},
year = {2015},
publisher = {Routledge},
doi = {10.1080/07294360.2014.934328}
}

@article{panadero2019using,
  title={Using formative assessment to influence self-and co-regulated learning: The role of evaluative judgement},
  author={Panadero, Ernesto and Broadbent, Jaclyn and Boud, David and Lodge, Jason M},
  journal={European Journal of Psychology of Education},
  volume={34},
  number={3},
  pages={535--557},
  year={2019},
  publisher={Springer}
}

@article{carless2023teacher,
  title={Teacher feedback literacy and its interplay with student feedback literacy},
  author={Carless, David and Winstone, Naomi},
  journal={Teaching in higher education},
  volume={28},
  number={1},
  pages={150--163},
  year={2023},
  publisher={Taylor \& Francis}
}

@article{carless2022teacher,
  title={From teacher transmission of information to student feedback literacy: Activating the learner role in feedback processes},
  author={Carless, David},
  journal={Active learning in higher education},
  volume={23},
  number={2},
  pages={143--153},
  year={2022},
  publisher={Sage Publications Sage UK: London, England}
}

@article{chong2021reconsidering,
  title={Reconsidering student feedback literacy from an ecological perspective},
  author={Chong, Sin Wang},
  journal={Assessment \& Evaluation in Higher Education},
  volume={46},
  number={1},
  pages={92--104},
  year={2021},
  publisher={Taylor \& Francis}
}

@article{wood2023enabling,
  title={Enabling feedback seeking, agency and uptake through dialogic screencast feedback},
  author={Wood, James},
  journal={Assessment \& Evaluation in Higher Education},
  volume={48},
  number={4},
  pages={464--484},
  year={2023},
  publisher={Taylor \& Francis}
}

@article{ziqi2024l2,
  title={L2 students’ barriers in engaging with form and content-focused AI-generated feedback in revising their compositions},
  author={Ziqi, Chen and Xinhua, Zhu and Qi, Lu and Wei, Wei},
  journal={Computer Assisted Language Learning},
  pages={1--21},
  year={2024},
  publisher={Taylor \& Francis}
}

@article{er2021collaborative,
  title={A collaborative learning approach to dialogic peer feedback: a theoretical framework},
  author={Er, Erkan and Dimitriadis, Yannis and Ga{\v{s}}evi{\'c}, Dragan},
  journal={Assessment \& Evaluation in Higher Education},
  volume={46},
  number={4},
  pages={586--600},
  year={2021},
  publisher={Taylor \& Francis}
}

@article{tam2025interacting,
  title={Interacting with ChatGPT for internal feedback and factors affecting feedback quality},
  author={Tam, Angela Choi Fung},
  journal={Assessment \& Evaluation in Higher Education},
  volume={50},
  number={2},
  pages={219--235},
  year={2025},
  publisher={Taylor \& Francis}
}

@article{zou2026investigating,
  title={Investigating students’ uptake of teacher-and ChatGPT-generated feedback in EFL writing: A comparison study},
  author={Zou, Shaoyan and Guo, Kai and Wang, Jun and Liu, Yu},
  journal={Computer Assisted Language Learning},
  volume={39},
  number={4},
  pages={1062--1091},
  year={2026},
  publisher={Taylor \& Francis}
}

@article{qian2025pedagogical,
  title={Pedagogical applications of generative AI in higher education: A systematic review of the field},
  author={Qian, Yufeng},
  journal={TechTrends},
  pages={1--16},
  year={2025},
  publisher={Springer}
}

@article{kondo2026ai,
  title={AI-Generated Versus Human Supervisor Feedback on Medical Students’ Clinical Clerkship Logs: Cross-Sectional Convergent Mixed Methods Study},
  author={Kondo, Takeshi and Donkers, Jeroen and Nishigori, Hiroshi and Rovers, Sanne and Heeneman, Sylvia},
  journal={JMIR Medical Education},
  volume={12},
  pages={e90064},
  year={2026},
  publisher={JMIR Publications Toronto, Canada}
}

@article{brummernhenrich2025applying,
  title={Applying social cognition to feedback chatbots: Enhancing trustworthiness through politeness},
  author={Brummernhenrich, Benjamin and Paulus, Christian L and Jucks, Regina},
  journal={British Journal of Educational Technology},
  volume={56},
  number={6},
  pages={2321--2340},
  year={2025},
  publisher={Wiley Online Library}
}

@article{usher2025generative,
  title={Generative AI vs. instructor vs. peer assessments: A comparison of grading and feedback in higher education},
  author={Usher, Maya},
  journal={Assessment \& Evaluation in Higher Education},
  volume={50},
  number={6},
  pages={912--927},
  year={2025},
  publisher={Taylor \& Francis}
}

@article{gladovic2025feedback,
  title={The feedback process as a scaffold of evaluative judgement},
  author={Gladovic, Cedomir and Tai, Joanna Hong-Meng and Nicola-Richmond, Kelli and Dawson, Phillip},
  journal={Higher Education Research \& Development},
  pages={1--16},
  year={2025},
  publisher={Taylor \& Francis}
}

@article{malecka2022eliciting,
  title={Eliciting, processing and enacting feedback: mechanisms for embedding student feedback literacy within the curriculum},
  author={Malecka, Bianka and Boud, David and Carless, David},
  journal={Teaching in higher education},
  volume={27},
  number={7},
  pages={908--922},
  year={2022},
  publisher={Taylor \& Francis}
}

@article{wood2021dialogic,
  title={A dialogic technology-mediated model of feedback uptake and literacy},
  author={Wood, James},
  journal={Assessment \& Evaluation in Higher Education},
  volume={46},
  number={8},
  pages={1173--1190},
  year={2021},
  publisher={Taylor \& Francis}
}

@article{molloy2020developing,
  title={Developing a learning-centred framework for feedback literacy},
  author={Molloy, Elizabeth and Boud, David and Henderson, Michael},
  journal={Assessment \& Evaluation in Higher Education},
  volume={45},
  number={4},
  pages={527--540},
  year={2020},
  publisher={Taylor \& Francis}
}

@article{nazaretsky2026gives,
  title={Who Gives Feedback Matters: Student Biases Towards Human and AI-Generated Formative Feedback},
  author={Nazaretsky, Tanya and Mejia-Domenzain, Paola and Swamy, Vinitra and Frej, Jibril and K{\"a}ser, Tanja},
  journal={Journal of Computer Assisted Learning},
  volume={42},
  number={1},
  pages={e70153},
  year={2026},
  publisher={Wiley Online Library}
}

@article{NAZARETSKY2026100533,
title = {Can students judge like experts? A large-scale study on the pedagogical quality of AI and human personalized formative feedback},
journal = {Computers and Education: Artificial Intelligence},
volume = {10},
pages = {100533},
year = {2026},
issn = {2666-920X},
doi = {https://doi.org/10.1016/j.caeai.2025.100533},

author = {Tanya Nazaretsky and Hagit Gabbay and Tanja Käser}
}

@article{guardia2026human,
  title={Human and AI-generated feedback in higher education: A systematic review of effectiveness and student perceptions},
  author={Guardia-Paniura, Candy Haydee and Cueva-Luza, Timoteo and Cruz-Carpio, Favio Mauricio and Ito-D{\'\i}az, Ra{\'u}l Reynaldo and Apaza-Paco, David Victor and Rosas-Rojas, Nilda and Mamani-Mamani, Benedicta and Terrero-P{\'e}rez, {\'A}ngel and Yaed{\'u}, Renato Yassutaka Faria and Peralta-Mamani, Mariela},
  journal={Contemporary Educational Technology},
  volume={18},
  number={1},
  pages={ep623},
  year={2026},
  publisher={Bastas}
}

@article{winstone2022discipline,
  title={Discipline-specific feedback literacies: A framework for curriculum design},
  author={Winstone, Naomi E and Balloo, Kieran and Carless, David},
  journal={Higher Education},
  volume={83},
  number={1},
  pages={57--77},
  year={2022},
  publisher={Springer}
}

@article{winstone2017supporting,
  title={Supporting learners' agentic engagement with feedback: A systematic review and a taxonomy of recipience processes},
  author={Winstone, Naomi E and Nash, Robert A and Parker, Michael and Rowntree, James},
  journal={Educational psychologist},
  volume={52},
  number={1},
  pages={17--37},
  year={2017},
  publisher={Taylor \& Francis}
}

@article{hattie2007power,
  title={The power of feedback},
  author={Hattie, John and Timperley, Helen},
  journal={Review of educational research},
  volume={77},
  number={1},
  pages={81--112},
  year={2007},
  publisher={Sage Publications Sage CA: Thousand Oaks, CA}
}

@article{khosravi2019ripple,
  title={Ripple: A crowdsourced adaptive platform for recommendation of learning activities},
  author={Khosravi, Hassan and Kitto, Kirsty and Williams, Joseph Jay},
  journal={arXiv preprint arXiv:1910.05522},
  year={2019}
}

@article{wu2026different,
  title={Different Trajectories of Student Feedback Literacy Development Through Learning-Centered Feedback Process in University Writing Class},
  author={Wu, Peng and Jiang, Xia},
  journal={International Journal of Applied Linguistics},
  volume={36},
  number={2},
  pages={1345--1358},
  year={2026},
  publisher={Wiley Online Library}
}

@article{Winstone03072023,
author = {Naomi E. Winstone and Robert A. Nash},
title = {Toward a cohesive psychological science of effective feedback},
journal = {Educational Psychologist},
volume = {58},
number = {3},
pages = {111--129},
year = {2023},
publisher = {Routledge},
doi = {10.1080/00461520.2023.2224444}
}

@article{carless2018development,
  title={The development of student feedback literacy: Enabling uptake of feedback},
  author={Carless, David and Boud, David},
  journal={Assessment \& evaluation in higher education},
  volume={43},
  number={8},
  pages={1315--1325},
  year={2018},
  publisher={Taylor \& Francis}
}

@article{tai2018developing,
  title={Developing evaluative judgement: enabling students to make decisions about the quality of work},
  author={Tai, Joanna and Ajjawi, Rola and Boud, David and Dawson, Phillip and Panadero, Ernesto},
  journal={Higher education},
  volume={76},
  number={3},
  pages={467--481},
  year={2018},
  publisher={Springer}
}

@article{yan2025distinguishing,
  title={Distinguishing performance gains from learning when using generative AI},
  author={Yan, Lixiang and Greiff, Samuel and Lodge, Jason M and Ga{\v{s}}evi{\'c}, Dragan},
  journal={Nature Reviews Psychology},
  volume={4},
  number={7},
  pages={435--436},
  year={2025},
  publisher={Nature Publishing Group US New York}
}

@article{bearman2024developing,
  title={Developing evaluative judgement for a time of generative artificial intelligence},
  author={Bearman, Margaret and Tai, Joanna and Dawson, Phillip and Boud, David and Ajjawi, Rola},
  journal={Assessment \& Evaluation in Higher Education},
  volume={49},
  number={6},
  pages={893--905},
  year={2024},
  publisher={Taylor \& Francis}
}

@article{jin2025students,
  title={Students’ perceptions of generative AI--powered learning analytics in the feedback process: A feedback literacy perspective},
  author={Jin, Flora Ji-Yoon and Maheshi, Bhagya and Lai, Wenhua and Li, Yuheng and Gasevic, Danijela and Chen, Guanliang and Charwat, Nicola and Chan, Philip Wing Keung and Martinez-Maldonado, Roberto and Ga{\v{s}}evi{\'c}, Dragan and others},
  journal={Journal of Learning Analytics},
  volume={12},
  number={1},
  pages={152--168},
  year={2025},
  publisher={Society for Learning Analytics Research}
}

@article{Little02012024,
author = {Tegan Little and Phillip Dawson and David Boud and Joanna Tai},
title = {Can students’ feedback literacy be improved? A scoping review of interventions},
journal = {Assessment \& Evaluation in Higher Education},
volume = {49},
number = {1},
pages = {39--52},
year = {2024},
publisher = {Routledge},
doi = {10.1080/02602938.2023.2177613}

}

@article{moore2024harnessing,
  title={Harnessing generative AI (GenAI) for automated feedback in higher education: A systematic review},
  author={Moore, Robert L and Lee, Sophia S},
  journal={Online Learning Journal},
  year={2024}
}

@article{chan2024generative,
  title={Generative AI and Essay Writing: Impacts of Automated Feedback on Revision Performance and Engagement.},
  author={Chan, Sumie and Lo, Noble and Wong, Alan},
  journal={Reflections},
  volume={31},
  number={3},
  pages={1249--1284},
  year={2024},
  publisher={ERIC}
}

@inproceedings{nedrehagen2025scoping,
  title={A scoping review of dialogic formative feedback practices in higher education},
  author={Nedrehagen, Eirik and Vee, Tove Sandvoll and Stokstad, Janette Moland and Orm, Stian},
  booktitle={Frontiers in Education},
  volume={10},
  pages={1696703},
  year={2025},
  organization={Frontiers Media SA}
}

@article{pitt2023enabling,
  title={Enabling and valuing feedback literacies},
  author={Pitt, Edd and Winstone, Naomi},
  journal={Assessment \& Evaluation in Higher Education},
  volume={48},
  number={2},
  pages={149--157},
  year={2023},
  publisher={Taylor \& Francis}
}

@article{myers2025dialogism,
  title={Dialogism in feedback literacies: a critical review},
  author={Myers, Tony and Buchanan, Jaime},
  journal={Assessment \& Evaluation in Higher Education},
  volume={50},
  number={6},
  pages={846--860},
  year={2025},
  publisher={Taylor \& Francis}
}

@article{ilangakoon2022relationship,
  title={The relationship between feedback and evaluative judgement in undergraduate nursing and midwifery education: An integrative review},
  author={Ilangakoon, Chanika and Ajjawi, Rola and Endacott, Ruth and Rees, Charlotte E},
  journal={Nurse Education in Practice},
  volume={58},
  pages={103255},
  year={2022},
  publisher={Elsevier}
}

@article{nicol2024shifting,
  title={Shifting feedback agency to students by having them write their own feedback comments},
  author={Nicol, David and Kushwah, Lovleen},
  journal={Assessment \& Evaluation in Higher Education},
  volume={49},
  number={3},
  pages={419--439},
  year={2024},
  publisher={Taylor \& Francis}
}

@article{yang2025feedback,
  title={Feedback Literacy and EFL Learner Engagement With ChatGPT Feedback: Predicting Feedback Uptake and Perceived Usefulness.},
  author={Yang, Yang and Hussin, Supyan and Hashim, Harwati},
  journal={Theory \& Practice in Language Studies (TPLS)},
  volume={15},
  number={11},
  year={2025}
}

@article{karunarathne2024evaluating,
  title={Evaluating student feedback literacies: a study using first-year business and economics students},
  author={Karunarathne, Wasana and Selman, Chris and Ryan, Tracii},
  journal={Assessment \& Evaluation in Higher Education},
  volume={49},
  number={4},
  pages={471--484},
  year={2024},
  publisher={Taylor \& Francis}
}

@article{ajjawi2018examining,
author = {Rola Ajjawi and David Boud},
title = {Examining the nature and effects of feedback dialogue},
journal = {Assessment \& Evaluation in Higher Education},
volume = {43},
number = {7},
pages = {1106--1119},
year = {2018},
publisher = {Routledge},
doi = {10.1080/02602938.2018.1434128}



}

@incollection{ajjawi2018conceptualising,
  title={Conceptualising evaluative judgement for sustainable assessment in higher education},
  author={Ajjawi, Rola and Tai, Joanna and Dawson, Phillip and Boud, David},
  booktitle={Developing evaluative judgement in higher education},
  pages={7--17},
  year={2018},
  publisher={Routledge}
}

@article{boud2013rethinking,
  title={Rethinking models of feedback for learning: the challenge of design},
  author={Boud, David and Molloy, Elizabeth},
  journal={Assessment \& Evaluation in Higher Education},
  volume={38},
  number={6},
  pages={698--712},
  year={2013},
  publisher={Taylor \& Francis}
}

@article{zimmerman2002becoming,
  title={Becoming a self-regulated learner: An overview},
  author={Zimmerman, Barry J},
  journal={Theory into practice},
  volume={41},
  number={2},
  pages={64--70},
  year={2002},
  publisher={Taylor \& Francis}
}

@article{nicol2006formative,
  author  = {Nicol, David J. and Macfarlane-Dick, Debra},
  title   = {Formative Assessment and Self-Regulated Learning: A Model and Seven Principles of Good Feedback Practice},
  journal = {Studies in Higher Education},
  year    = {2006},
  volume  = {31},
  number  = {2},
  pages   = {199--218},
  doi     = {10.1080/03075070600572090}
}

@article{cong2025critical,
  title={Critical thinking in the age of generative AI: Effects of a short-term experiential learning intervention on EFL learners},
  author={Cong-Lem, Ngo and Nguyen, Thang Tat and Nguyen, Khanh Nhat Hoang},
  journal={International Journal of TESOL Studies},
  volume={250522},
  pages={1--21},
  year={2025}
}

@article{hon2026generative,
  title={Generative AI in higher education: A systematic review of its effects on learning outcomes and academic performance},
  author={Hon, Kellie},
  journal={Journal of Educational Technology Systems},
  volume={54},
  number={3},
  pages={537--560},
  year={2026},
  publisher={SAGE Publications Sage CA: Los Angeles, CA}
}

@article{zhan2025generative,
  title={Generative artificial intelligence as an enabler of student feedback engagement: a framework},
  author={Zhan, Ying and Boud, David and Dawson, Phillip and Yan, Zi},
  journal={Higher Education Research \& Development},
  volume={44},
  number={5},
  pages={1289--1304},
  year={2025},
  publisher={Taylor \& Francis}
}

@article{paris2022instructors,
  title={Instructors’ perspectives of challenges and barriers to providing effective feedback},
  author={Paris, Brit and others},
  journal={Teaching and Learning Inquiry},
  volume={10},
  year={2022}
}

@article{haughney2020quality,
  title={Quality of feedback in higher education: A review of literature},
  author={Haughney, Kathryn and Wakeman, Shawnee and Hart, Laura},
  journal={Education Sciences},
  volume={10},
  number={3},
  pages={60},
  year={2020},
  publisher={MDPI}
}

@incollection{esterhazy2020counts,
  title={What counts as quality feedback? Disciplinary differences in students’ and teachers’ perceptions of feedback},
  author={Esterhazy, Rachelle and Fossland, Trine and Stalheim, Odd-Rune},
  booktitle={Quality work in higher education: Organisational and pedagogical dimensions},
  pages={155--174},
  year={2020},
  publisher={Springer}
}

@article{deneen2023connecting,
  title={Connecting teacher and student assessment literacy with self-evaluation and peer feedback},
  author={Deneen, Christopher C and Hoo, Hui-Teng},
  journal={Assessment \& Evaluation in Higher Education},
  volume={48},
  number={2},
  pages={214--226},
  year={2023},
  publisher={Taylor \& Francis}
}

@book{winstone2019designing,
  author    = {Winstone, Naomi and Carless, David},
  title     = {Designing Effective Feedback Processes in Higher Education: 
               A Learning-Focused Approach},
  publisher = {Routledge},
  year      = {2019},
  address   = {Abingdon},
  doi       = {10.4324/9781351115940}
}

@article{liu2025ai,
  title={AI feedback literacy in higher education: understanding, measuring, and predicting student feedback uptake},
  author={Liu, Ke and Deris, Farhana Diana},
  journal={Assessment \& Evaluation in Higher Education},
  pages={1--15},
  year={2025},
  publisher={Taylor \& Francis}
}

@article{banihashem2024feedback,
  author    = {Banihashem, Seyyed Kazem and Kerman, Nafiseh Taghizadeh 
               and Noroozi, Omid and Moon, Jon and Drachsler, Hendrik},
  title     = {Feedback Sources in Essay Writing: Peer-Generated or 
               {AI}-Generated Feedback?},
  journal   = {International Journal of Educational Technology in 
               Higher Education},
  year      = {2024},
  volume    = {21},
  number    = {1},
  pages     = {23},
  doi       = {10.1186/s41239-024-00455-4}
}

@article{MUNOZMUNOZ2025103805,
title = {ChatGPT-generated versus human direct corrective feedback on L2 writing},
journal = {System},
volume = {134},
pages = {103805},
year = {2025},
issn = {0346-251X},
doi = {https://doi.org/10.1016/j.system.2025.103805},

author = {Belén C. {Muñoz Muñoz} and Hossein Nassaji and Felipe I. {Bello Carrillo}}
}

@inproceedings{alnemrat2025ai,
  title={AI vs. teacher feedback on EFL argumentative writing: a quantitative study},
  author={Alnemrat, Areen and Aldamen, Hesham and Almashour, Mohamad and Al-Deaibes, Mutasim and AlSharefeen, Rami},
  booktitle={Frontiers in Education},
  volume={10},
  pages={1614673},
  year={2025},
  organization={Frontiers Media SA}
}

@article{pozdniakov2026,
title = {AI assistance in peer feedback provision: Pedagogically sound, but minimally adopted},
journal = {Computers \& Education},
volume = {248},
pages = {105591},
year = {2026},
issn = {0360-1315},
doi = {https://doi.org/10.1016/j.compedu.2026.105591},

author = {Stanislav Pozdniakov and Jonathan Brazil and Seyyed Kazem Banihashem and Omid Noroozi and Dragan Gašević and Shazia Sadiq and Hassan Khosravi}
}

@article{er2025assessing,
  author    = {Er, Erkan and Noroozi, Omid and Banihashem, Seyyed Kazem},
  title     = {Assessing Student Perceptions and Use of Instructor Versus 
               {AI}-Generated Feedback},
  journal   = {British Journal of Educational Technology},
  year      = {2025},
  volume    = {56},
  pages     = {1074--1091},
  doi       = {10.1111/bjet.13558}
}

@article{guo2025peer,
  author    = {Guo, Kai and Zhang, E. and Li, Danling and Yu, Shulin},
  title     = {Using {AI}-supported peer review to enhance feedback 
               literacy: An investigation of students' revision of 
               feedback on peers' essays},
  journal   = {British Journal of Educational Technology},
  year      = {2025},
  volume    = {56},
  number    = {4},
  pages     = {1612--1639},
  doi       = {10.1111/bjet.13540}
}

\end{document}